\documentclass[11 pt, oneside]{article}

\usepackage[title]{appendix}
\usepackage[pdftex]{graphicx}
\usepackage{hyperref}
\usepackage {subcaption}
\usepackage{mathtools}
\usepackage{setspace}
\usepackage[normalem]{ulem}
\usepackage{float}
\usepackage[paperwidth=8.5in, paperheight=11in, top=1in, bottom=1in, left=1in, right=1in]{geometry}

\usepackage{endnotes}
\let\footnote=\endnote

\usepackage{abstract}
\usepackage{geometry}
\usepackage{graphicx}
\usepackage{setspace}
\usepackage{natbib}

\usepackage[scaled=0.9]{helvet}

\usepackage{booktabs}
\usepackage {amsbsy}
\usepackage {amssymb}
\usepackage {amsmath}
\usepackage{amsthm}
\usepackage{multirow}
\usepackage{algorithmic}
\usepackage{algorithm}
\usepackage{enumerate}
\usepackage[affil-sl,auth-lg]{authblk}

\usepackage{bbm}
\usepackage{natbib}
 \bibpunct{(}{)}{;}{a}{,}{,}
\setcitestyle{citesep={,}}

\usepackage{soul} % for strikethrough
\usepackage{color}
\newcommand{\alert}{\textcolor{black}}
\newcommand{\comm}{\textcolor{black}}
\newcommand{\BFcomm}{\textcolor{black}}
\newcommand{\apcomm}{\textcolor{black}}

\newcommand{\BFcommTwo}{\textcolor{black}}
\newcommand{\apcommtwo}{\textcolor{black}}
\newcommand{\BFcommThree}{\textcolor{black}}

\DeclareMathOperator*{\argmin}{arg\,min}
\DeclareMathOperator*{\diag}{diag}
\newcommand{\BTheta}{\boldsymbol{\Theta}}
\newcommand{\Btheta}{\boldsymbol{\theta}}

\newcommand{\Bpsi}{\boldsymbol{\psi}}

\newcommand{\rl}{\mathbb{R}}

\newcommand{\Bnu}{\boldsymbol{\nu}} 
\newcommand{\Btau}{\boldsymbol{\tau}}
\newcommand{\BEll}{\boldsymbol{\ell}}
\newcommand{\BSigma}{\boldsymbol{\Sigma}}

\newcommand{\BI}{\mathbf{I}}
\newcommand{\BV}{\mathbf{V}}
\newcommand{\Bv}{\mathbf{v}}

\newcommand{\Ncal}{\mathcal{N}}
\theoremstyle{plain}

\newtheorem{proposition}{Proposition}
\newtheorem{lem}{Lemma}
\theoremstyle{remark}
\newtheorem{rmk}{Remark}
\newcommand{\norm}[1]{\left\lVert#1\right\rVert}
\newcommand{\suchthat}{\;\ifnum\currentgrouptype=16 \middle\fi|\;}

\date{}
\begin{document}

\title{A Bayesian framework for functional calibration of expensive computational models through non-isometric matching}

\author{Babak Farmanesh, Arash Pourhabib, Balabhaskar Balasundaram, Austin Buchanan\thanks{School of Industrial Engineering and Management, Oklahoma State University, \{babak.farmanesh, arash.pourhabib, baski, buchanan\}@okstate.edu}}
\maketitle
\vspace{-.5cm}
\begin{abstract}
\noindent\textbf{\textit{Abstract}}: We study statistical calibration, i.e., adjusting features of a computational model that are not observable or controllable in its associated physical system. We focus on functional calibration, which arises in many manufacturing processes where the unobservable features, called calibration variables, are a function of the input variables. A major challenge in many applications is that computational models are expensive and can only be evaluated a limited number of times. Furthermore, without making strong assumptions, the calibration variables are not identifiable. We propose Bayesian non-isometric matching calibration (BNMC) that allows calibration of expensive computational models with only a limited number of samples taken from a computational model and its associated physical system. BNMC replaces the computational model with a dynamic Gaussian process (GP) whose parameters are trained in the calibration procedure. To resolve the identifiability issue, we  present the calibration problem from a geometric perspective of non-isometric curve to surface matching, which enables us to take advantage of combinatorial optimization techniques to extract necessary information for constructing prior distributions. Our numerical experiments demonstrate that in terms of prediction accuracy BNMC outperforms, or is comparable to, other existing calibration frameworks. 
 
\end{abstract}
{\bf Keywords}: Functional calibration, Gaussian process, Generalized minimum spanning tree.

\section {Introduction}
\label{sec.intro}
Experimenting on \BFcommThree{computational models} to understand physical systems has been a popular practice ever since computers became advanced enough to handle complex mathematical
models and intense computational procedures~\citep{fang2005design,santner2013design}. This popularity is mainly because a computational model can obtain outputs of an experiment in a relatively more cost-effective and timely manner compared to conducting actual experiments in a laboratory. However, one challenge in utilizing computational models is their ``adjustment." In fact, computational models usually incorporate features that cannot be observed or measured in physical systems, but must be correctly specified so that the computational model can accurately represent the physical system~\citep{kennedy2001bayesian}. We refer to these unobservable/unmeasurable features as \emph{calibration variables}, and to the adjustment of their values as the \emph{calibration procedure}. We call the input features which are common between the computational models and the physical systems as \emph{control variables}.

For example, in the fabrication of poly-vinyl alcohol (PVA) treated buckypaper, we are interested in understanding the relationship between the
response value, which is the tensile strength, and the control variable, which is the PVA
amount~\citep{PFC}. Here, the calibration variable is the percentage of
PVA absorbed, which cannot be measured in the physical system, but is required in the computational model.

Past studies on the calibration problem generally assumed unique values for calibration variables, an approach referred to as \emph{global calibration}, and used different statistical approaches to estimate these values. For instance,~\citet{kennedy2001bayesian,craig2001bayesian,reese2004integrated,
higdon2004combining,higdon2008computer,
williams2006combining,bayarri2007framework}, and~\citet{goldstein2009reified} devised various Bayesian models, whereas~\cite{loeppky2006computer} and~\cite{pratola2013fast} used maximum likelihood estimation, and~\cite{joseph2009statistical} and~\citet{han2009simultaneous} developed mixed models by combining frequentist and Bayesian methodologies. More recently~\cite{tuo2015efficient, tuo2016theoretical} developed models based on $L_2$ distance projection to estimate the true values of the global calibration variables.

Presently, few studies employ \emph{functional calibration} by assuming that the values of the calibration variables \emph{depend on the control variables}.~\cite{PFC} showed that, for the buckypaper fabrication problem, an approach that considers a parametric functional relationship between the amount of PVA and the percentage absorbed can outperform the global calibration approach of~\cite{kennedy2001bayesian}. Similarly,~\cite{xiong2009better} used a simple linear relationship to improve the calibration accuracy in a benchmark thermal challenge problem. \apcomm{Furthermore, non-parametric methods can also be utilized to model functional relationships between the calibration variables and the control variables.} Such non-parametric functional relationships have been constructed using Reproducing Kernel Hilbert Spaces~\citep{scholkopf2001generalized} and Gaussian processes~\citep{rasmussen2004gaussian} by various authors~\citep{NFC,plumlee2016calibrating,NBC}.

All the aforementioned studies in functional calibration and most  studies in global calibration require computational models that are ``cheaply executable.'' This assumption is required since computational models need to be evaluated thousands of times either to draw samples from a posterior distribution in Bayesian approaches, or to numerically minimize a loss function in other approaches. If the computational model is ``expensive,'' one can obtain a small number of observations from the computational model, then fit a surrogate function based on these random samples, and in the final step, replace the computational model in the calibration procedure with this new surrogate model. However, as discussed in Section~\ref{sec.results}, this poses a challenge because ``static'' replacement may result in poor retrieval of the calibration variables.

%However, surrogate modeling itself creates an additional source of error due to restrictions in drawing a sufficient number of random samples from the expensive computational model, which results in deterioration of the calibration accuracy.
Another challenge is the identifiability issue: it is difficult to solve the calibration problem in higher dimensional spaces without making additional assumptions about the solution space~\citep{NFC}. Furthermore, good prediction performance for the response values does not necessarily imply that a method has accurately captured the functional relationship between the calibration and control variables\BFcomm{~\citep{ tuo2015efficient, plumlee2016orthogonal, ezzat2018sequential}}. This is a significant drawback since, in many applications, understanding the functional relationship between the calibration and control variables is as important as predicting the response values of the system under study. 

In this paper, we develop a new framework for the functional calibration of expensive computational models. Unlike conventional surrogate modeling, which replaces the computational model with a static, approximated surface, we employ a ``dynamic'' Gaussian process (GP) over the computational model. Our GP is dynamic in the sense that the hyper-parameters of the GP's covariance function are trained during the calibration procedure. We simultaneously construct posterior distributions for the hyper-parameters of the GP's covariance function and the calibration variables associated with each of the physical control vectors. In other words, we allow the GP to tune its hyper-parameters in addition to the calibration variables such that the computational model responses become as close as possible to the physical responses.

%In fact, the focus of local calibration approaches of~\cite{NBC} and~\cite{NFC} are on the accurate estimation of the functional relationship between the control and the calibration variables. Accurate estimation however, requires the evaluation of the computational models a large number of times, which may not be feasible for  expensive computational models. This results in a poor retrieval of the relationship between the control and the calibration variables. Therefore, in our approach, we compensate this inaccurate retrieval by allowing the GP to tune its hyper-parameters along with the calibration variables in a way that the computational model responses become as close as possible to the physical responses.

\begin{figure}[H]
\begin{center}
	\begin{subfigure}{0.48\textwidth}
		\includegraphics[height=6cm,width=7.5cm]{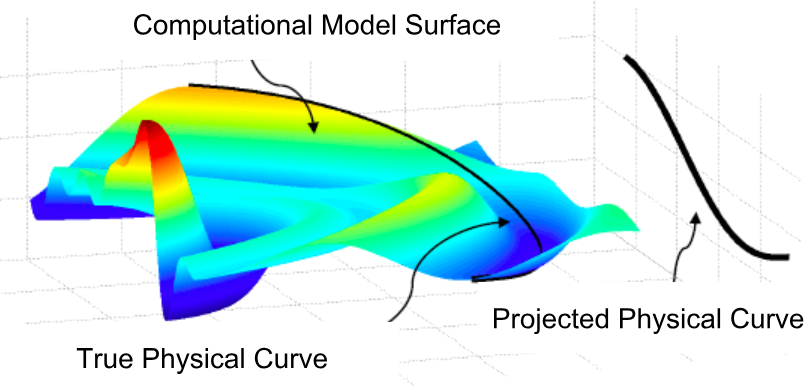}
		\caption {Complete curve and surface}
\label{fig:CurveToSurfaceComp}
	\end{subfigure}
	\begin{subfigure}{0.48\textwidth}
		\includegraphics[height=6cm,width=7.5cm]{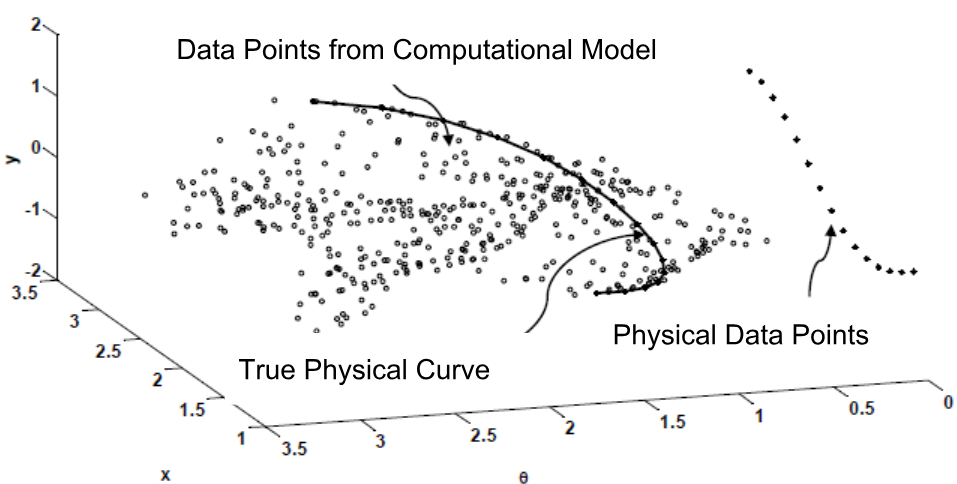}
		\caption {Observed curve and surface by incomplete data }
\label{fig:CurveToSurfaceScatter}
	\end{subfigure}
\end{center}
\caption{A non-isometric curve to surface matching perspective of functional calibration: The left plot shows the  complete  surface and curve. In practice, we observe a scatter of data points sampled from the complete curve and surface, which is depicted in the right plot.}
\label{fig:CurveToSurface}
\end{figure}

To tackle the unidentifiability issue in higher-dimensional spaces, we use informative prior distributions. We take advantage of an alternative geometric interpretation of calibration, namely the non-isometric matching of a curve to a surface. We explain this in the case of a single control variable and a single calibration variable. From a geometric perspective, all possible values for the control variable and the physical response constitute a plane curve in the control-response space (see Figure~\ref{fig:CurveToSurfaceComp}). By contrast, in the computational model, we can specify the values of both the control and the calibration variables. Consequently, all possible values of the control and calibration variables, and the responses of the computational model together form a surface.  The  plane physical curve we observe in the control-response space is a projection of a space curve in the three-dimensional control-calibration-response space. By nature of projection into a lower-dimensional space, the length of the projected curve is not necessarily the same as the original curve in the three-dimensional space. The projection is therefore \textit{non-isometric}\BFcommTwo{~\citep{bronstein2003expression, bronstein2005three}.}

The geometric interpretation is due to the nature of the calibration variable in a physical process: for each value of the control variable there exists a (possibly unknown) value for the calibration variable, and these two features determine a single response. Since we do not observe the actual value of the calibration variable in the
physical process, we only see a projected curve in the control-response space. Hence, calibration aims to recover the true physical curve, or in other words, determine a non-isometric
match of a curve to a surface.

The remainder of this paper is organized as follows. We explain our Bayesian model for handling expensive computational models in Section~\ref{sec.BayesianModel}. Section~\ref{sec.CurveToSurfaceMatching} contains a formal description of the calibration problem and its interpretation as a non-isometric curve to surface matching problem. Our graph-theoretic approach to utilizing this geometric perspective to construct informative prior distributions for our Bayesian model is also presented in Section~\ref{sec.CurveToSurfaceMatching}. In Section~\ref{sec.CalibrationGraph}, we generalize the idea of a non-isometric curve to surface matching to higher dimensions and introduce integer programming techniques to tackle the problem.  The approach presented in Sections~\ref{sec.CurveToSurfaceMatching} and~\ref{sec.CalibrationGraph} to construct informative prior distributions for our Bayesian model is used to calculate posterior distributions in Section~\ref{sec.priors}.  Our experimental results are reported in Section~\ref{sec.results}, comparing them with previous approaches. Section~\ref{sec.concl} concludes the paper and presents paths for future research.

\section{General setting: a Bayesian model for calibration}\label{sec.BayesianModel}
Consider a physical system that operates according to a set of (possibly unknown) physical laws. In this system, there is a functional relationship between a group of features  and the response (output). We call those features of the system that can be measured and specified as inputs of the physical system as \emph{control variables}, and denote the vector of these variables by $\mathbf{x} \in \rl^{d^x}$. We assume that we obtain data for the physical system by conducting \emph{physical experiments}: once  the control variables are set (either observed or specified) in the physical system, the physical process $\mathcal{F}^p$ generates a real-valued response $y^p$, that is $y^p=\mathcal{F}^p(\mathbf{x})$.

Although the response is a function of \textit{all} features of the physical system, we write $y^p$ explicitly as a function of  $\mathbf{x}$ as the rest of the features are hard to measure or control, and hence we have no control over them in the physical system. We call such features \emph{calibration variables} and categorize them into the following two groups:~(i) \emph{global calibration variables}, which have unique values regardless of the values of the control variables, and~(ii) \emph{functional calibration variables}, which are functions of control variables. 

We denote the vector of global calibration variables by $\Bpsi \in \rl^{d^\psi}$ and the vector of functional calibration variables by $\Btheta \in \rl^{d^\theta}$. We also denote the function that maps $\mathbf{x}$ to the $k^\text{th}$ element of $\Btheta$, i.e., $\theta_k$, by $\mathcal{F}^\theta_k$ and the vector of all these functions by $\mathcal{F}^{\theta}=[\mathcal{F}^{\theta}_1,\ldots,\mathcal{F}^{\theta}_{d^{\theta}}]^{\top}$. With a slight abuse of notation, we denote the vector map from $\mathbf{x}$ to $\Btheta $ using the vector of functions $\mathcal{F}^{\theta}$ as $\Btheta=\mathcal{F}^\theta(\mathbf{x})$. 
%Finally, given a positive integer  $n$, we compactly denote the set of all positive integers up to $n$ as $[n] \coloneqq \{1,\ldots,n\}$.

Suppose we have a computational model  constructed according to the laws governing the physical system. Similar to the physical system, the response of our computational model is determined by the interactions between the control and the calibration variables. However, in a computational model we can set the values of all $\mathbf{x}$, $\Btheta $, and $\Bpsi $ arbitrarily within their respective domains. That is because, unlike physical experiments, there are no constraints on measuring or specifying control or calibration variables in a \BFcommThree{computational} model. If we denote the computational process as $\mathcal{F}^s$, then the  response of the  computational model  can be written as,
\begin{equation}
y^s= \mathcal{F}^s(\mathbf{x},\Bpsi, \Btheta ).\label{ComputationalModel}
\end{equation}
We refer to obtaining a value for $y^s$, given a combination of $\mathbf{x}$, $\Bpsi$, and $\Btheta$ in the computational model, as a \emph{computer experiment}.

The goal of \emph{calibration} is to adjust the variables $\Bpsi\text{ and }\Btheta$ such that the computational model represents the physical system in the sense that the computational model can predict the physical response at any input location $\mathbf{x}^*$.

Mathematically, calibration can be viewed as the estimation of vectors $\mathcal{F}^{\theta} \text{ and } \Bpsi$ such that, for any given $\mathbf{x}^{*}$, the function $\mathcal{F}^s:\rl^{d^x}\times \rl^{d^\psi} \times \rl^{d^\theta}\longrightarrow \rl$ generates a response close to $y^{p^*}$ up to an error $\epsilon^{*}$, i.e., 
\apcommtwo{
\begin{eqnarray}
y^{p^*}= \mathcal{F}^s(\mathbf{x}^*,\boldsymbol{\psi}, \mathcal{F}^{\theta}(\mathbf{x}^*))+\epsilon^*,\label{ComPhyRel}
\end{eqnarray}
where the error $\epsilon^*$ captures the measurement error and the discrepancy between the physical and computation model.}
%\begin{eqnarray}
%y^{p^*}= \mathcal{F}^s(\mathbf{x}^*,\boldsymbol{\tilde%{\psi}}, \mathcal{\tilde{F}}^{\theta}(\mathbf{x}^*))+\epsilon^*,\label{ComPhyRel}
%\end{eqnarray}
%where $\boldsymbol{\tilde{\psi}}$ and $ \mathcal{\tilde{F}}^{\theta}$ are \BFcomm{estimates} of $\mathcal{F}^{\theta}$ and $ \Bpsi$, and the error $\epsilon^*$ exists due to assumptions and simplifications made in the computational model and also due to the estimation of the calibration variables. 

To estimate $\boldsymbol{\psi}$ and $ \mathcal{F}^{\theta}$ in~\eqref{ComPhyRel} we initially obtain $m$ responses from $\mathcal{F}^p$ at a set of physical system inputs $\{\mathbf{x}^p_1,\ldots,\mathbf{x}^p_m\}$ to create a dataset $P$ corresponding to that physical system, \[P \coloneqq \left\{p_i=\left(\mathbf{x}^p_i,y^p_i\right) \suchthat \mathbf{x}_i^p\in \rl^{d^x}, y_i^p\in \rl, i \in \{1,2,\ldots,m\}\right\}.\] 
We also create the counterpart of $P$ in the computational model, i.e., the  computational dataset as \[S \coloneqq \left\{s_j=\left(\mathbf{x}^s_j,\Bpsi_j^s,\Btheta^s_j,y^s_j\right) \suchthat \mathbf{x}_j^s\in \rl^{d^x},\Bpsi^s_j\in \rl^{d^\psi}, \Btheta^s_j\in \rl^{d^\theta},j \in \{1,2,\ldots,n\}\right\},\] based on a set of computational model inputs  $\{(\mathbf{x}^s_1,\Bpsi^s_1,\Btheta^s_1),\ldots,(\mathbf{x}^s_n,\Bpsi^s_n,\Btheta^s_n)\}$. \apcomm{We assume the sets of physical system inputs and the computational model inputs are given. For a discussion of how to select the inputs we refer the reader to the paper by~\citet{ezzat2018sequential}}.

\BFcommTwo{Let $\Btheta^p_i=\mathcal{F}^\theta(\mathbf{x}^p_i)$ and $\Bpsi^p$ denote the true values of the calibration variables and assume that the errors are independent and have identical normal distribution with zero mean and constant variance. Therefore, we obtain the calibration model as  }
\begin{eqnarray}
y^p_i= \mathcal{F}^s(\mathbf{x}_i^p,\Bpsi^p, \Btheta^p_i)+\epsilon_i^p, \text{ where } \epsilon_i^p\sim \mathcal{N}(0,\sigma^2),\quad \forall i\in\{1,2,\ldots,m\}.\label{NonVecModel}
\end{eqnarray}

\begin{rmk}
%If we remove the functional calibration variable $\Btheta^p_i$ from equation~\eqref{NonVecModel}, we get a simplified version of the global calibration model proposed in~\citep{kennedy2001bayesian}. In fact,~\cite{kennedy2001bayesian} assume $y^p_i= \mathcal{F}^s(\mathbf{x}_i^p,\Bpsi^p)+\delta(\mathbf{x}_i^p)+\epsilon_i^p$, where $\delta(\cdot)$ is a GP independent from $ \mathcal{F}^s$, which characterizes all the discrepancy between the computational model and the physical system due to assumptions made in building the computational model. However, because we use a dynamic GP to minimize the discrepancy, we choose to use $\epsilon_i^p$ to represent not only the measurement error in the physical system but also the discrepancy between the computational model and the physical system. \apcommtwo{As such, our assumptions are similar to those made by~\cite{NBC}. In Appendix~\ref{Res_analysis} we validate these assumptions on the datasets used in this study.} The reader can refer to the discussion by~\cite{tuo2016theoretical} for a frequentist interpretation of the model proposed by~\cite{kennedy2001bayesian}.

If we remove the functional calibration variable $\Btheta^p_i$ from equation~\eqref{NonVecModel}, we get a simplified version of the global calibration model proposed in~\citep{kennedy2001bayesian}. In fact,~\cite{kennedy2001bayesian} assume $y^p_i= \mathcal{F}^s(\mathbf{x}_i^p,\Bpsi^p)+\delta(\mathbf{x}_i^p)+\epsilon_i^p$, where $\delta(\cdot)$ is a GP independent from $ \mathcal{F}^s$, which characterizes all the discrepancy between the computational model and the physical system due to assumptions made in building the computational model. \BFcommThree{ However, because in this study we focus on computational models with a limited number of data points, we do not include a separate discrepancy term  $\delta(\cdot)$, which entails introducing a set of additional parameters and would make the estimation procedure unstable. Therefore, as we utilize a dynamic GP to minimize the overall discrepancy, we choose to use $\epsilon_i^p$ to represent not only the measurement error in the physical system but also the discrepancy between the computational model and the physical system}. As such, our assumptions are similar to those made by~\cite{NBC}. In Appendix~E we validate these assumptions on the datasets used in this study. The reader can refer to the discussion by~\cite{tuo2016theoretical} for a frequentist interpretation of the model proposed by~\cite{kennedy2001bayesian}.

\end{rmk}

Note that applying Bayesian statistics to construct posterior distributions for parameters of model~\eqref{NonVecModel}, i.e., $\Bpsi^p, \sigma^2$ and $\Btheta^p_i$, requires a large number of evaluations of  $\mathcal{F}^s$, which is not practical for expensive computational models. Therefore, we assume $\mathcal{F}^s$ is a GP \apcommtwo{with a constant mean}~\citep{rasmussen2004gaussian}, i.e.,  $\mathcal{F}^s \sim \mathcal{GP}(0,\mathcal{K(\cdot,\cdot)})$, where $\mathcal{K(\cdot,\cdot)}$ is a covariance function. \apcommtwo{We further assume that the overall average of the responses from the computational model has been subtracted from each output, and as such we use a GP with mean zero.} Here we use the squared exponential kernel function as the choice of the covariance function,
\begin{eqnarray}
\mathcal{K}(\mathbf{z},\mathbf{z'})=\gamma\exp(-(\mathbf{z}-\mathbf{z'})^{\top} \mathbf{L}(\mathbf{z}-\mathbf{z'})),\label{SqrExpKernel}
\end{eqnarray}
where $\gamma$ is the magnitude parameter and $\mathbf{L}$ is a diagonal matrix of the length-scale parameters. We denote the vector of the diagonal elements of $\mathbf{L}$ by $\BEll$. 

Subsequently, we can obtain the likelihood of model~\eqref{NonVecModel} by the GP distribution defined on $\mathcal{F}^s$ as a multivariate normal distribution,
\begin{eqnarray}
\mathbf{y}^p\suchthat\mathbf{X}^p,\BTheta^p,\Bpsi^p,\BEll,\gamma,\sigma^2 \sim \mathcal{N}(0,\BSigma +\sigma^2 \BI_m),\label{likelihood}
\end{eqnarray}
where $\mathbf{y}^p=[y^p_1,\ldots,y^p_m]^{\top}$ is the vector of physical responses, $\mathbf{X}^p=[\mathbf{x}_1^p,\ldots,\mathbf{x}_m^p]^{\top}$ and  $\BTheta^p=[\Btheta^p_1,\ldots,\Btheta^p_m]^{\top}$ are matrices of size $m \times d^x$ and $m \times d^\theta$ respectively, and $\BSigma$ is the $m\times m$ covariance matrix whose elements are calculated by covariance function~\eqref{SqrExpKernel} with $[\mathbf{x}^{p^{\top}}_i,\Btheta^{p^{\top}}_i,\Bpsi^{p^{\top}}]^{\top}$ as input vectors with length $(d^x+d^{\theta}+d^{\psi})$.

\begin{rmk}
Although in the process of deriving likelihood~\eqref{likelihood}, we consider $\Bpsi^p$ and the columns of $\BTheta^p$ as the input variables of model~\eqref{NonVecModel}, we do not know the values of these input variables, and we intend to estimate them. Therefore, in order to distinguish the calibration variables $\Bpsi$ and $\Btheta$, in~\eqref{ComputationalModel} from the parameters in model~\eqref{NonVecModel}, we refer to $\BTheta^p$ and $\Bpsi^p$ as \emph{calibration parameters}.
\end{rmk}

We can estimate the calibration parameters of model~\eqref{NonVecModel}, the parameters of covariance function~\eqref{SqrExpKernel}, and the variance of error, using Bayesian statistics with the posterior distribution,
\begin{eqnarray}
\pi(\BTheta^p,\Bpsi^p,\BEll,\gamma,\sigma^2\suchthat\mathbf{y}^p,\mathbf{X}^p)\propto \pi(\mathbf{y}^p\suchthat\mathbf{X}^p,\BTheta^p,\Bpsi^p,\BEll,\gamma,\sigma^2)\pi(\BTheta^p)\pi(\Bpsi^p)\pi(\BEll)\pi(\gamma)\pi(\sigma^2). \label{posterior_Init}
\end{eqnarray}
% where we substitute $\BPsi^p$ by $\Bpsi^p$, since all columns of $\BPsi^p$ are essentially the same and $\BPsi^p$ can be easily  recovered  by the following transformation: $\BPsi^p=\mathbbm{1}_{m\times d_\psi} \diag(\Bpsi^p)^T$, where $\mathbbm{1}_{m\times d_\psi}$ is a $m\times d_\psi$ matrix of ones. 

The Bayesian model~\eqref{posterior_Init} would be completed by specifying prior distributions for the parameters $\pi(\BTheta^p)$, $\pi(\Bpsi^p)$, $\pi(\BEll)$, $\pi(\gamma)$, and $\pi(\sigma^2)$. However, our model suffers from unidentifiablity in the absence of informative priors due to the high-dimensionality of the parameter space. \apcomm{Therefore, in Section~\ref{sec.priors} we present graph-theoretic approaches that help construct informative priors for the calibration parameters $\BTheta^p$ and $\Bpsi^p$}.

Note that the replacement of $\mathcal{F}^s$ by $ \mathcal{GP}(0,\mathcal{K(\cdot,\cdot)})$ does not constitute a surrogate modeling approach, wherein the computational model is replaced by a \emph{fixed} surrogate surface, which is in turn trained based on a set of limited samples drawn from the computational model prior to any calibration procedure. Our approach is fundamentally different from surrogate modeling, since building and training the GP is a part of the calibration process.

\section{Calibration as non-isometric matching: a special case}\label{sec.CurveToSurfaceMatching}

We explain in this section how the calibration problem can be viewed as a non-isometric curve to surface matching problem for the special case where $\mathbf{x}\in \rl$, $\Btheta\in \rl$, and  $\Bpsi\in \varnothing$, which means both control and calibration variables are one-dimensional and no global calibration variable exists. From a geometric perspective, all the possible
values for $\mathbf{x}$ and $\mathcal{F}^p(\mathbf{x})$ constitute the curve $(\mathbf{x},\mathcal{F}^p(\mathbf{x}))$ in a two-dimensional space. In the computational model, however, we can specify the values of both $\mathbf{x}$ and $ \Btheta$. Consequently, all the possible values
of $\mathbf{x}$, $ \Btheta$, and $\mathcal{F}^s(\mathbf{x},\Btheta)$ together form a surface $(\mathbf{x}, \Btheta,\mathcal{F}^s(\mathbf{x},\Btheta))$ in a three-dimensional space. 
As we noted in Section~\ref{sec.intro}, the true physical curve lies on the three-dimensional computational model surface, i.e., $(\textbf{x},\mathcal{F}^{\theta}(\mathbf{x}),\mathcal{F}^p(\mathbf{x}))$. However, since we do not observe the actual values of the
calibration variables in the physical process, we only see a projected
curve in $\mathbf{x}-y$ space (see Figure~\eqref{fig:CurveToSurfaceComp}). Hence, the calibration problem is to
recover the true physical curve, or, in other words, determine a non-isometric
match of a curve to a surface. 

As mentioned earlier, the non-isometry is due to the fact that the curve $(\textbf{x},\mathcal{F}^{\theta}(\mathbf{x}),\mathcal{F}^p(\mathbf{x}))$ on the three-dimensional $\mathbf{x}-  \Btheta-y$ space has a different length than the projected curve $(\mathbf{x},0,\mathcal{F}^p(\mathbf{x}))$ on a two-dimensional $\mathbf{x}-y$ space. Therefore, this is, in
principle, different from isometric matching problems~\citep{gruen2005least,bronstein2005three,baltsavias2008high}.

In practice we only have the finite physical system dataset $P$ along with a finite computational model dataset $S$, as we do not observe a complete curve or surface. Ideally, the points in $P$ lie on the projected curve that we observe, and the points in $S$ lie on the computational model surface (see Figure~\ref{fig:CurveToSurfaceScatter}). Hence, what we observe is
incomplete data, and we aim to match non-isometrically an
incomplete curve to an incomplete surface, which is equivalent to solving the calibration problem.

This geometric perspective motivates us to view the problem through a combinatorial lens and model the problem using graph-theoretic approaches. Our graph-based solution to the non-isometric curve to surface matching problem provides us with a set of computational model data points, which carry information about the calibration parameters. We call this set of computational data points \emph{anchor points}. These anchor points will then be used in Section~\ref{sec.priors} to construct prior distributions for our Bayesian model.

We seek to identify a set of anchor points among the computational data points that are ``close'' to the points on the true physical curve. In other words, the anchor points are positioned such that the true physical curve passes through the neighborhoods of those points. We want the anchor points to satisfy two desirable properties: (i) the computational model response should be close to the physical response for a given input $\mathbf{x}$; and (ii) the calibration parameter values for two consecutive anchor points should be close to each other. The former drives our method to identify the anchor points that have similar responses to that of the physical system, and the latter aims to encourage the smoothness of the  physical curve. 

Note that we are only interested in identifying these ``optimal'' anchor points that provide us with information about the true physical curve to be used in our prior distributions, and not  the true physical curve itself. However, one could also directly use the selected anchor points to approximate the true physical curve via interpolation. Given our focus on expensive computational models wherein the number of computational model data points is limited, such an approximation of the true physical curve may not be accurate. In the next section, we formally define and address the problem of finding anchor points with the desired properties using a graph-theoretic approach \apcommtwo{for the special case when  $\mathbf{x}\in \rl$, $\Btheta\in \rl$, and $\Bpsi\in \varnothing$.}

\subsection{A graph-theoretic approach for finding anchor points}\label{ssec.DAGSPPmodel}

Without loss of generality, we assume that all the data points in the physical system and the computational model datasets are strictly ordered such that $\mathbf{x}^p_i<\mathbf{x}^p_{i+1}$, for all $i\in \{1,2,\ldots,m-1\}$, and $\mathbf{x}^s_j<\mathbf{x}^s_{j+1}$, for all $j\in \{1,2,\ldots,n-1\}$.  \comm {We construct an edge-weighted directed graph  $G=(V,E)$ with vertex set  $V \coloneqq   \{0,1,2,\ldots,n+1\}$, and the edge set $E$ described in equation~\eqref{eq:edgedefin} below. The vertices in  $V^0  \coloneqq \{1,2,\ldots,n\}$ correspond to the computational model data points in $S$.} We refer to $G$ as the \emph{calibration digraph}.

\begin{figure}
\centering
\includegraphics[width=\textwidth]{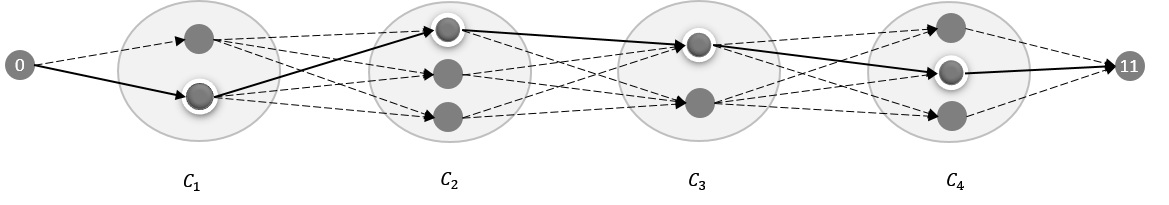}
%\vspace{-150pt}
\caption{Illustration of the calibration digraph for the case where $m=4$ and $n=10$. The vertices represent data points from the computational model and the clusters $C_1$ through $C_m$ correspond to physical system data points. Vertices denoted by dark circles with a white border represent the anchor points, and the solid arrows identify the edges in the shortest path found. }
\label{fig_graph}
\end{figure}

Recall that, intuitively, the objective of calibration is to minimize the difference between the outputs of the physical system and the corresponding computational model. As such, the first step is to find control variables that are similar in both settings, the physical experiments and the computer experiments. Therefore, we first group the control variables in the computational model based on their distance to the control variables in the physical experiments. We partition $V^0$ into $m$ clusters $C_1,\ldots,C_{m}$ as follows:  any vertex $j \in V^0$ corresponding to data point $s_j \in S$ is assigned to a unique cluster $C_i$ by the following formula: 
\begin{equation}\label{eq:cluster_rule}
j \in C_i  \iff i  = \min\left\{ \argmin_{\ell \in \{1,2,\ldots,m\}}\{\vert\vert \mathbf{x}^p_\ell - \mathbf{x}^s_j\vert\vert_2\} \right\}.
\end{equation}
If the inner minimum in~\eqref{eq:cluster_rule} is not unique, then the outer minimum is used to break the tie by choosing the smallest index. As a consequence, each cluster $C_i$ is in 1-to-1 correspondence with the $i^{th}$ physical data point. \comm{This choice of tie-breaker is easy to implement and  establishes a mechanism for consistent assignment of points to a cluster.}
We can now describe the set of directed edges $E$ as
\begin{equation}\label{eq:edgedefin}
E \coloneqq \bigcup_{i = 1}^{m-1} \{(u,v) \suchthat u \in C_i, v \in C_{i+1}\} \bigcup \{(0,u) \suchthat u \in C_1\} \bigcup \{(u,n+1) \suchthat u \in C_m\}.
\end{equation}
This construction is illustrated in Figure~\ref{fig_graph}.

The final critical step is to assign a weight $w_{uv}$ to each edge $(u,v) \in E$. Consider two consecutive clusters $C_i$ and $C_{i+1}$  and vertices $u \in C_i$ and $v\in C_{i+1}$. Define $w_{uv}$ as
\begin{equation}\label{eq:edgewt}
w_{uv} \coloneqq  \vert y^s_u-y^p_u\vert +   \lambda \vert\vert\Btheta^s_u - \Btheta^s_v\vert\vert_2,
\end{equation}
where $\lambda > 0 $ is a scaling parameter. The weights of edges that leave vertex 0 or enter vertex $n+1$ are identically zero. The edge-weight for any edge between two consecutive clusters $i$ and $i+1$ consists of two parts: the first part $\vert y^s_u-y^p_u\vert$ represents the difference between the model response and physical response; the second part $\vert\vert\Btheta^s_u - \Btheta^s_v\vert\vert_2$ represents the difference between the calibration parameters of $i$ and $i+1$.
On this digraph $G$ with the given edge-weights, we intend to solve the shortest path problem from origin vertex $0$ to destination vertex $n+1$. Every path from vertex 0 to vertex $n+1$ in $G$ has exactly $m+1$ edges by construction. Suppose $0$-$v_1$-$v_2$-$\cdots$-$v_m$-$(n+1)$ is the shortest path identified. Then, those points in $S$ corresponding to  $\{v_1,\ldots, v_m\}$  serve as the anchor points. \apcomm{The edge-weights quantify the proximity of the physical and computer experiment outputs and difference between the calibration parameters to minimize erratic changes.}

\begin{lem}\label{lemma_dag}
Calibration digraph $G=(V,E)$ is acyclic with a topological ordering $\langle 0,1,\ldots,n,n+1\rangle$. 
\end{lem}

\apcomm{\begin{proof}
See Appendix~B for proofs.
\end{proof}}

Since $G$ is a directed acyclic graph, \apcomm{or DAG for short}, we can solve the shortest path problem using an $\mathcal{O}(|E|)$  algorithm that scans outgoing edges from each vertex in the topological order and updates distance-labels as needed~\citep{bellman1958routing,Lawler76book}. 

%\begin{algorithm}[H]
%\begin{algorithmic}[1]
%\STATE $dist(0):= 0, dist(u):=\infty\ \forall \ u \in V\setminus \{0\}$
%\STATE $pred(0):= 0, pred(u):= -1 \forall \ u \in V\setminus \{0\}$
%\FOR{$u=0,\ldots,n$}
%\FOR{each $(u,v) \in E$ }
%\IF{$dist(v) > dist(u)+w_{uv}$}
%\STATE $dist(v):= dist(u)+w_{uv}$
%\STATE $pred(v):= u$
%\ENDIF
%\ENDFOR
%\ENDFOR
%\end{algorithmic}
%\caption{Reaching algorithm for DAGs}\label{alg:reaching}
%\end{algorithm}

\section{Generalization of non-isometric matching to higher dimensions }\label{sec.CalibrationGraph}
Section~\ref{sec.CurveToSurfaceMatching} introduced the curve to surface matching interpretation of calibration with $\mathbf{x}\in \rl$ and $\Btheta \in \rl$. This special case allowed us to develop a graph-theoretic approach for anchor point selection that admitted a \comm{fast $\mathcal{O}(|E|)$  algorithm}.  The geometric perspective can be generalized to arbitrary dimensions as a  hyper-curve to hyper-surface matching problem.  However, in the general setting, there is no straightforward extension of the directed acyclic graph model. \apcommtwo{Recall that the model hinges on the natural ordering of the computational and the physical data points on the real line, which does not exist in higher dimensions. So, in this section we introduce a different  calibration graph model and an associated combinatorial optimization problem to find the anchor points in an arbitrary dimension.} As with the special case, the anchor points will subsequently be used in Section~\ref{sec.priors} to construct prior distributions for our Bayesian model.

\begin{figure}[H]
\begin{center}
	\begin{subfigure}{0.45\textwidth}
		\includegraphics[height=4cm,width=7cm]{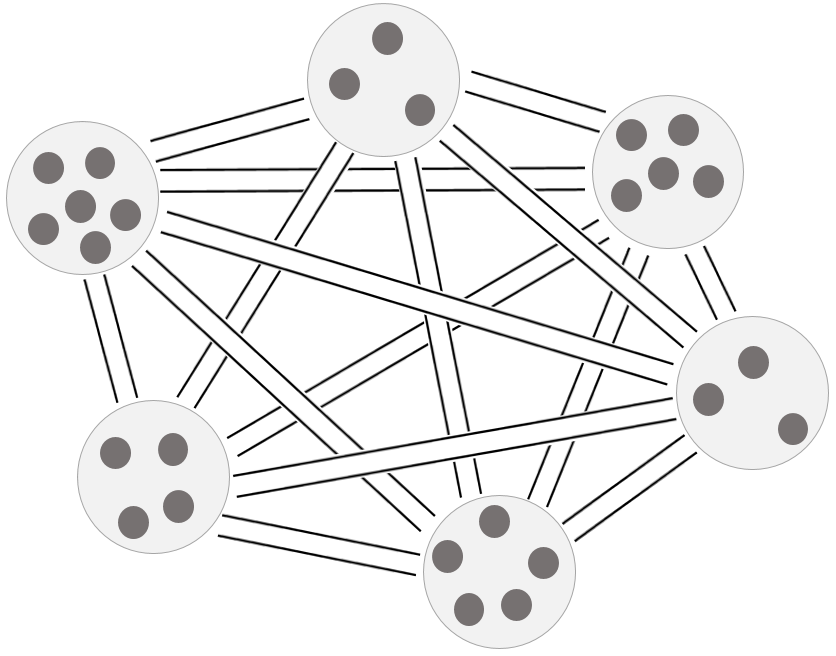}
		\caption {Calibration graph}
\label{fig:CalibrationNet}
	\end{subfigure}
	\begin{subfigure}{0.45\textwidth}
		\includegraphics[height=4cm,width=7cm]{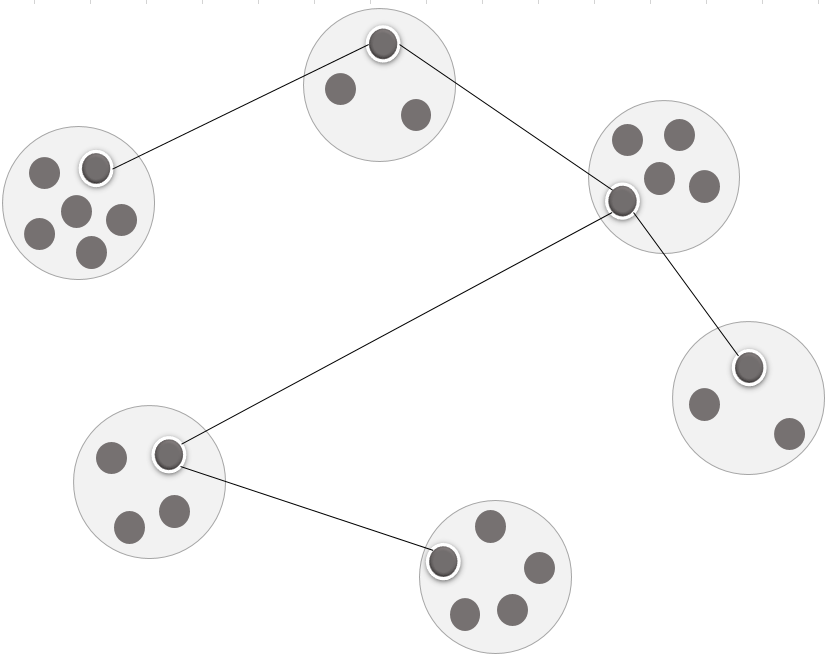}
		\caption {A generalized minimum spanning tree}
\label{fig:tree}
	\end{subfigure}
\end{center}
\caption{(a) A calibration graph where each black circle represents a vertex and each two parallel lines represent edges between vertices of two clusters. (b) A generalized spanning tree in the calibration graph.}
\label{fig:NetworkRep}
\end{figure}

For the general case, we construct a \emph{calibration graph} $G=(V,E)$ that is undirected and edge-weighted, where $V=\{1,2,\ldots,n\}$ corresponds to the $n$ computational data points. We partition $V$ into $m$ clusters, $C_1,\ldots,C_m$, in correspondence with the $m$ physical data points and assign vertex $j$ to a  cluster $C_i$ by the same rule in \comm{equation}~\eqref{eq:cluster_rule}. The graph $G$ is a complete $m$-partite graph with partitions $C_1,\ldots,C_m$, \comm{i.e., distinct vertices are adjacent if and only if they belong to  different partitions. The edge set can be described formally as} $E \coloneqq \bigcup_{i=1}^{m-1}\bigcup_{\ell=i+1}^m\left\{\{u,v\} \suchthat u\in C_i,v\in C_{\ell}\right\}.$ Figure~\ref{fig:CalibrationNet} illustrates this construction.

%\alert{\sout{We call two computational data points, $s_u, s_{v}\in S$, \emph{neighbors} if the Euclidean distance between their control vectors is smaller than a predefined radius $r$, i.e., $||\mathbf{x}^s_u-\mathbf{x}^s_{v}||_2<r$.}} 

Finally, before defining the edge weights, we introduce two required concepts. The \emph{calibration vector} of data point $s_j$ is given by $[\Btheta^{s^{\top}}_{j},\Bpsi^{s^{\top}}_{j}]^{\top}$. We assign the weight $w_e$ to the edge $e = \{u,v\}\in E$, where $u \in C_i$ and $v \in C_{\ell}$, by
\begin{equation}
w_e \coloneqq \left\{\begin{array}{cc}
 |y_u^s-y_i^p|+|y_{v}^s-y_{\ell}^p|+\lambda||[\Btheta^{s^{\top}}_u,\Bpsi^{s^{\top}}_u]^{\top}-[\Btheta^{s^{\top}}_{v},\Bpsi^{s^{\top}}_{v}]^{\top}||_2& \text{if }||\mathbf{x}^s_u-\mathbf{x}^s_{v}||_2 \leq r\\
|y_u^s-y_i^p|+|y_{v}^s-y_{\ell}^p|+M||\mathbf{x}^s_u-\mathbf{x}^s_{v}||_2& \text{if } ||\mathbf{x}^s_u-\mathbf{x}^s_{v}||_2>r,
\end{array} \right. 
\label{Genweight}
\end{equation}
where $\lambda$ is a scaling parameter and $M$ is a sufficiently large number used to penalize the computational data points that are far from each other. Note that the weights assigned in~\eqref{Genweight} extend the idea behind equation~\eqref{eq:edgewt}. Here, the edge weight between vertices $u \in C_i$ and $v \in C_{\ell}$, where $s_u$ and $s_v$ are \emph{neighbors} (\apcomm{that is, the Euclidean distance between their control vectors is smaller than a predefined radius $r$}), consists of two parts, similar to~\eqref{eq:edgewt}: the first part measures the distance between each vertex's response and the physical system response associated with the cluster to which it belongs, i.e., $|y_u^s-y_i^p|$ and $|y_{v}^s-y_{\ell}^p|$; the second part measures the distance between the corresponding calibration vectors, i.e., $||[\Btheta^{s^{\top}}_u,\Bpsi^{s^{\top}}_u]^{\top}-[\Btheta^{s^{\top}}_{v},\Bpsi^{s^{\top}}_{v}]^{\top}||_2$.  \comm{Let $E_1$ denote the set of all edges that join vertex pairs  corresponding to control vectors that are at most Euclidean distance $r$ apart.} The remainder of the edges, $E_2=E\setminus E_1$, correspond to edges between computational data points that \comm{are not close enough}, and we assign relatively large weights to these edges by setting $M$ to a large value. Furthermore, the weight on such edges increases as the distance between the control vectors of the end points increases.

To identify the ``optimal'' anchor vertices from this calibration graph, we find a minimum weight tree that contains exactly one vertex from each cluster. In the  optimization literature, this problem  is known as the \emph{generalized minimum spanning tree} (GMST) problem~\citep{myung1995generalized}. By our construction of the edge weights, a GMST will tend to include edges in $E_1$ as they are lighter. However, if no GMST exists that only uses edges in $E_1$, it will be forced to include edges in $E_2$.

\subsection{Integer programming \comm{approaches} to the GMST problem}\label{ssec.GMSTIP}
The GMST problem was introduced by~\cite{myung1995generalized},  who showed that it is NP-hard and does not admit a polynomial-time constant-factor approximation algorithm unless $\text{P} = \text{NP}$. Various authors have developed and analyzed integer programming (IP) formulations for this problem and the strength of the associated linear programming (LP) relaxations~\citep{myung1995generalized,feremans2002comparative,Pop2004, Pop2006}. \alert{Strong formulations, which correspond to tight LP relaxations, are desirable in a branch-and-bound algorithm as they produce tighter bounds that can be helpful in pruning the search tree.} \comm{We employ two such strong formulations with tight  LP relaxations for solving the anchor point selection problem in arbitrary dimension.} 

\comm{The class of formulations that were first introduced by~\cite{myung1995generalized} employs exponentially many constraints and  are analogous to the cutset and subtour elimination formulations of the traveling salesman problem and the minimum spanning tree problem~\citep{BertsimasIPbook}.    \cite{feremans2002comparative} showed that strengthening a subtour elimination formulation of a more general variant of the GMST problem
 is among the strongest in terms of the tightness of the LP relaxation.} \alert{We use this formulation, which \cite{feremans2002comparative} call the \emph{directed cluster subpacking} (DCSUB) formulation, in our computational experiments. This formulation and additional explanation are provided in Appendix~C.}

\comm{Because of the presence of exponentially many  constraints, a direct implementation of the \alert{entire} DCSUB formulation  is impractical even for small scale problems. Nonetheless, a delayed constraint generation approach could be effective in practice~\citep{Buchanan2015CDS,BBEMkclubCHC2018,BBYLEMkclub2018CHCerr}. This approach starts by relaxing the formulation  by omitting a subset of the constraints (typically those that are exponentially many in number). During the normal progress of an LP relaxation based branch-and-bound algorithm to solve the relaxed IP, whenever an integral solution is  detected at some node of the search tree, it is necessary to verify if a   constraint that violates this solution exists among those constraints that were excluded. If so, we solve the model at that node again after adding the violated constraints back; otherwise, we continue to branch as usual,} thus ensuring the overall correctness of the algorithm. An effective implementation of such an algorithm is possible using the ``lazy cut'' feature available in most state-of-the-art  IP solvers as long as the \comm{identification} of the violated constraints can be accomplished quickly.

The second type of formulation we use in our computational experiments is  based \comm{on the classical multi-commodity network flow (MCF) formulation}~\citep{myung1995generalized,feremans2002comparative,Pop2004,Pop2006}. \comm{The underlying idea of this formulation is to use the flow of a (dummy) ``commodity'' in the network to trace a path between two vertices by designating one vertex with unit supply for that commodity and the other with unit demand. As the MCF formulation  uses only polynomially many constraints and variables, it can be directly implemented and solved using most IP solvers for moderately sized instances.} \alert{The MCF formulation, which also has a strong LP relaxation, is presented and discussed in greater detail in Appendix~C.} 

\apcommtwo{Before proceeding to the next section, we point out that although GMST can be applied to any dataset (one- or multi-dimensional), the shortest path model is preferable for one-dimensional problems for the following reasons. First, the shortest path model does not require specifying parameters $r$ and $M$, and second, it is solvable in time $\mathcal{O}(|V|+|E|)$. By contrast, GMST requires choosing parameters $r$ and $M$, and it is NP-hard in general. For large scale one-dimensional problems, solving the GMST model may require a sophisticated integer programming approach and may require substantially more computational effort when compared to the shortest path alternative.}

%\BFcomm{Before, finishing this section, we acknowledge that translating the calibration as a case of non-isometric matching problem and subsequently as combinatorial optimization problems are the main contributions of this paper in Sections 3 and 4, but not solving the optimization problems.}

%For a detailed review of the IP formulations available for the GMST problem and a comparison of the strength of their LP relaxations, we refer the reader to the analysis by~\cite{feremans2002comparative}. 

\section{Prior and posterior distributions}\label{sec.priors}
This section describes how the information about the true physical curve carried by the anchor points, found by approaches discussed in Sections~\ref{sec.CurveToSurfaceMatching} and~\ref{sec.CalibrationGraph}, can be used to construct our prior distributions for the calibration parameters. We also expand posterior distribution~\eqref{posterior_Init} using the priors specified in this section, and show how we can make predictions at a new control vector $\mathbf{x}^*$.

Suppose $\Btheta^{a}_i$ and $\Bpsi^{a}_i$ are, respectively, the functional and the global calibration vectors of the anchor point associated with the $i^{th}$ physical data point, i.e., the anchor vertex selected from the $i^{th}$ cluster. \apcomm{Recall that the anchor points are selected by minimizing a weighted combination of two measures: a) the difference between the model responses and physical responses, and b) the distance between the corresponding calibration vectors. As such we can utilize those anchor points to build priors for the calibration parameters in a Bayesian model. To account for the uncertainty associated with the selection of the anchor points, we use variance hyperparameters as explained next}. Define the matrix $\BTheta^{a}\coloneqq [\Btheta^{a}_1,\ldots,\Btheta^{a}_m]^{\top}$  of size $m \times d^{\theta}$ and the mean vector $\Bpsi^a \coloneqq \frac{1}{m}\sum_{i=1}^{m}\Bpsi^a_i$ of length $d^\psi$. Note that for the latter we take the average of the global calibration vectors of the anchor points since we assume that the global calibration parameters are constant regardless of the values of the control vectors.

For each component of $\Bpsi^{p}$, \apcomm{the global calibration parameters}, we consider a univariate normal distribution centered at the corresponding element in $\Bpsi^{a}$ with an unknown variance as the choice of the prior distribution. Therefore, we construct the prior distribution for $\Bpsi^p$ as
\begin{eqnarray}
\Bpsi^p\suchthat \Bpsi^a,\Btau^2 \sim \mathcal{N}
(\Bpsi^a, \diag(\Btau^2)),\label{psiprior}
\end{eqnarray}
where $\Btau^2=[\tau^2_1,\ldots,\tau^2_{d^\psi}]^{\top}$ is the vector of variances of the normal distribution.

Applying the same procedure for constructing prior distributions for the functional calibration parameters increases the dimension of the parameter space, since we need to define $md^\theta$ variance parameters, which are nuisance parameters and not of interest to our model. Therefore, in order to shrink the parameter space, we use the fact that the $k^{th}$ column of \apcomm{the functional calibration parameters} $\BTheta^p$, i.e. $\BTheta^p_k$,  is actually a realization of the functional relationship $\mathcal{F}^{\theta}_k$. Therefore, the $k^{th}$ column of  $\BTheta^a$, i.e., $\BTheta^a_k$, is a rough estimator of this realization. On this basis, we use a single variance parameter for all the elements in $\BTheta^p_k$, and construct the prior distribution for $\BTheta^{p}_k$ as
\begin{eqnarray}
\BTheta^{p}_k\suchthat\BTheta^{a}_k,\nu_k^2~\sim \mathcal{N}(\BTheta^{a}_k,\nu_k^2\BI_m),\label{thataprior}
\end{eqnarray}
where $\nu_k^2$ is the $k^{th}$ element of the vector of variances $\Bnu^2$ with length $d^{\theta}$.

The correctness of the normality assumptions in~\eqref{psiprior} and~\eqref{thataprior} is a legitimate concern, because there is no guarantee that the anchor points embrace the true physical curve due to the limited number of observations. However, we only make the normality assumptions in~\eqref{psiprior} and~\eqref{thataprior} for constructing the prior distributions, and the Bayesian model will adjust these priors by likelihood~\eqref{likelihood}.

To specify the posterior distribution, we define proper prior distributions for the rest of the parameters. As such, we get:
\begin{eqnarray}
\begin{split}
\pi(\BTheta^p,\Bpsi^p,\Bnu^2,\Btau^2,\BEll,\gamma,\sigma^2\suchthat\mathbf{y}^p,\mathbf{X}^p,\BTheta^{a},\Bpsi^a)\propto\\
\pi(\mathbf{y}^p\suchthat\mathbf{X}^p,\BTheta^p,\Bpsi^p,\BEll,\gamma,\sigma^2)\pi(\BTheta^p\suchthat\BTheta^a,\Bnu^2)\pi(\Bpsi^p\suchthat\Bpsi^a,\Btau^2)\pi(\Bnu^2)\pi(\Btau^2)\pi(\BEll)\pi(\gamma)\pi(\sigma^2),\label{posterior_concise}
\end{split}
\end{eqnarray}
where,
\begin{eqnarray*} 
\begin{aligned}
& \mathbf{y}^p\suchthat\mathbf{X}^p,\BTheta^p,\Bpsi^p,\BEll,\gamma,\sigma^2 \sim \mathcal{N}(0,\BSigma +\sigma^2 \BI_m).\\
%& \BTheta^{p}_k\suchthat\BTheta^{a}_k,{\nu}_k^2\ \sim \mathcal{N}(\BTheta^{a}_k,\nu^2_k \BI_m),&\forall k\in [d^{\theta}],\\
%&\Bpsi^p\suchthat\Bpsi^a,\Btau^2\sim \mathcal{N}(\Bpsi^a, \diag(\Btau^2)),\\
%& \nu^2_k \sim \frac{1}{\nu_k^2},&\forall k\in [d^{\theta}],\\
%& \tau^2_h \sim \text{Inv-Gamma} (\alpha_{\tau},\beta_{\tau}),&\forall h\in[d^{\psi}],\\
%& \ell_j\sim \text{Log-Gamma}(\alpha_{\ell},\beta_{\ell}), &\forall j\in [d^x+d^{\theta}+d^{\psi}],\\
%& \gamma\sim \text{Log-Uniform},\\
%& \sigma^2\sim \text{Log-Uniform}.\label{posterior3}
\end{aligned}
\end{eqnarray*} 
\apcomm{We refer the reader to Appendix~A for the exact prior distributions of parameters in~\eqref{posterior_concise}, and a discussion of how to  sample from the posterior distribution.}

In order to make predictions at a new control vector $\mathbf{x}^*$, we introduce variables
$\Bpsi^p(t)$, $\BEll(t)$, $\gamma(t)$, $\sigma^2(t)$, and $\BTheta^p(t)$ as $t^{th}$ draws from the posterior distribution~\eqref{posterior_concise} after some burn-in period, where $t \in \{1,\ldots,T\}$. The first step in the prediction of response $y^*$ is to estimate the associated functional calibration vector of $\mathbf{x}^*$, i.e., $\Btheta^*=\mathcal{F}^{\theta}(\mathbf{x}^*)$. We can estimate $\Btheta^*$ based on each $\BTheta^p(t)$, where we denote the $t^{\text{th}}$ estimation of $\Btheta^*$ based on $\BTheta^p(t)$ as $\Btheta^*(t)$. To this end we note that as $\BTheta^p_k(t)$ is a vector of estimates of $\mathcal{F}^\theta_k$ at the design locations $\{\mathbf{x}^p_1,\ldots,\mathbf{x}^p_m\}$, we can write $\BTheta^{p}_k(t)=[\mathcal{F}^\theta_k(\mathbf{x}_1^p),\ldots,\mathcal{F}^\theta_k(\mathbf{x}^p_m)]^{\top}+[\epsilon^\theta_1,\ldots,\epsilon^\theta_m]^{\top}$, 
where $[\epsilon^\theta_1,\ldots,\epsilon^\theta_m]$ is a vector of the corresponding error terms. The error term appears because $\BTheta^p_k(t)$ does not contain exact evaluations of the function $\mathcal{F}^\theta_k$ but only estimations. \BFcomm{The following proposition obtains the mean prediction of $\theta_k^*(t)$, i.e., the $t^{th}$ estimation of $k^{th}$ element of $\Btheta^*$ base on $\BTheta_k^p(t)$.}

\begin{proposition}\label{Prop_ThetaStarPred}
Assume $[\epsilon^\theta_1,\ldots,\epsilon^\theta_m]^{\top}\sim \mathcal{N}(0,\sigma^\theta_k\BI_m)$ and $\mathcal{F}^\theta_k$ is a GP with mean zero and covariance function $\mathcal{K}$, i.e., $\mathcal{F}^\theta_k \sim \mathcal{GP}(0,\mathcal{K(\cdot,\cdot)})$, then  
\begin{eqnarray}
\theta_{k}^*(t)=\BSigma_{\mathbf{x}^*\mathbf{X}^p}(\BSigma_{\mathbf{X}^p\mathbf{X}^p}+\sigma^{\theta}_k\BI_m)^{-1}\BTheta^{p}_k(t).\label{PredTheta}
\end{eqnarray}
\end{proposition}
%\BFcomm{\begin{proof}
%See Appendix~\ref{Proof_Prop_ThetaStarPred}
%\end{proof}}

To find point and interval predictions for the new response $y^*$, we make $T$ predictions based on the $T$ samples we drew from the posterior~\eqref{posterior_concise} and the $T$ predictions we made for the vector $\Btheta^*$ using~\eqref{PredTheta}. Recall from Section~\ref{sec.BayesianModel} that $\mathcal{F}^s \sim \mathcal{GP}(0,\mathcal{K(\cdot,\cdot)})$; therefore, we can
use the GP predictive distribution to derive the $t^{th}$ prediction as
\begin{eqnarray}
\mathcal{F}^s(\mathbf{x}^*,\Btheta^*(t),\Bpsi^p(t))\sim \Ncal\bigg(\BSigma_{\Bv^*(t)\BV(t)}\big(\BSigma_{\BV(t)\BV(t)}+\sigma^2(t)\BI_m\big)^{-1}\mathbf{y}^p,\nonumber\\
\BSigma_{\Bv^*(t)\Bv^*(t)}-\BSigma_{\Bv^*(t)\BV(t)}\big(\BSigma_{\BV(t)\BV(t)}+\sigma^2(t)\BI_m\big)^{-1}\BSigma_{\BV(t)\Bv^*(t)}\bigg),\label{PredDistFor1Sample}
\end{eqnarray}
where $\Bv^*(t)=[\mathbf{x}^{*^{\top}},\Btheta^{*^{\top}}(t),\Bpsi^{p^{\top}}(t)]^{\top}$, $\BV(t)=[\mathbf{X}^P,\BTheta^P(t),\mathbbm{1}_{m\times d_\psi}\text{diag}(\Bpsi^p(t))]^{\top}$, and the covariance matrices are calculated using the $t^{th}$ sample of the covariance parameters, namely $\BEll(t)$ and $\gamma(t)$.

Finally, we derive our prediction using distribution~\eqref{PredDistFor1Sample} as
\begin{eqnarray*}
\hat{\mu}^*=\frac{1}{T} \sum_{t=1}^{T}\bigg(\BSigma_{\Bv^*(t)\BV(t)}\big(\BSigma_{\BV(t)\BV(t)}+\sigma^2(t)\BI_m\big)^{-1}\mathbf{y}^p\bigg),\\
\hat{\sigma}^{*^2}=\frac{1}{T^2} \sum_{t=1}^{T}\bigg(\BSigma_{\Bv^*(t)\Bv^*(t)}-\BSigma_{\Bv^*(t)\BV(t)}\big(\BSigma_{\BV(t)\BV(t)}+\sigma^2(t)\BI_m\big)^{-1}\BSigma_{\BV(t)\Bv^*(t)}\bigg).
\end{eqnarray*}

\section{Experimental results}\label{sec.results}
In this section, we evaluate the performance of our methodology by testing it on three synthetic problems and two real problems. We use the root mean squared error (RMSE) as the measure of accuracy in prediction of responses and calibration vectors to compare the performance of the competing methodologies;
\begin{eqnarray*}
\text{RMSE}_y=\sqrt{\frac{1}{n^*} \sum_{q=1}^{n^*}(\hat{y}_q^*-y_q^*)^2} &\text{and}&
\text{RMSE}_\theta=\sqrt{\frac{1}{n^*} \sum_{q=1}^{n^*}\norm{[\hat \Btheta^{*^{\top}}_q,\hat \Bpsi^{*^{\top}}_q]^{\top}-[\bar \Btheta^{*^{\top}}_q,\bar \Bpsi^{*^{\top}}_q]^{\top}}_2^2},
\end{eqnarray*}
where $\hat{y_q}^*$, $\hat{\Btheta}_q^*$, and $\hat{\Bpsi}_q^*$ are the predicted response and calibration parameter values for the $q^{\text{th}}$ test control variable.

Both the MCF and DCSUB formulations find anchor points for each dataset in less than two minutes on all the instances in our testbed, which shows that both formulations are fast in terms of computation time. To choose proper values of $\lambda$, $r$, and $M$ for the GMST model, we use the following empirical approach. First, in order to have the same scale for the inputs and outputs in $S$ and $P$, before constructing the calibration graph we standardize (divide by the range of each element) the control set $\{\mathbf{x}^p_i,\mathbf{x}^s_j\suchthat i\in\{1,2,\ldots,m\}, j\in \{1,2,\ldots,n\}\}$, the calibration set $\{\Btheta^s_j,\Bpsi^s_j\suchthat j\in \{1,2,\ldots,n\}\}$, and the response set $\{y^s_j,y^p_i\suchthat i\in\{1,2,\ldots,m\}, j\in \{1,2,\ldots,n\}\}$. We denote the standardized versions of $\mathbf{x}^p_i,\mathbf{x}^s_j,\Btheta^s_j,\Bpsi^s_j,y^s_j$, and $y^p_i$ by $\bar{\mathbf{x}}^p_i,\bar{\mathbf{x}}^s_j,\bar \Btheta^s_j,\bar \Bpsi^s_j,\bar y^s_j$, and $\bar y^p_i$. Note that this standardization is only for finding the anchor points, and once the anchor points are chosen, we transform the data back to their original scales for Bayesian inference. After standardization, $\lambda$ has to be less than two, otherwise the closeness of the calibration vectors is weighted as more important than closeness of the responses. We choose $\lambda$ from $\{0.1, 0.5, 1,1.5\}$ in our experiments. Moreover, for $M$ to be a sufficiently large number, it should be larger than the sum of all the weights on the arcs that can be potentially in $E_1$, that is:
\begin{eqnarray*}
M\geq \sum_{j\in\{1,2,\ldots,n\}}\sum_{k\in\{1,2,\ldots,n\}}\sum_{i\in\{1,2,\ldots,m\}}\sum_{\ell\in\{1,2,\ldots,m\}}|\bar y_j^s-\bar y_i^p|+|\bar y_k^s-\bar y_\ell^p|+\lambda \norm{[\bar \Btheta^{s^{\top}}_j,\bar \Bpsi^{s^{\top}}_j]^{\top}-[\bar \Btheta^{s^{\top}}_{k},\bar \Bpsi^{s^{\top}}_{k}]^{\top}}_2.
\end{eqnarray*} 
Finally, to choose a proper value for $r$, we  plot the pairwise Euclidean distances between all $\bar{\mathbf{x}}^p$ vectors. Then, we use the fact that each point on this plot represents an existing distance between the centers of two clusters in the calibration graph. Therefore, an upper bound on a group of smallest distances, which are close together and disjoint from other groups of distances, can be used as a proper value for $r$.

\subsection{Description of the calibration problems}\label{sec_Syn_Prob_Desc}
\apcomm{Table~\ref{table:SynProbsFunctions} describes synthetic problems used in this study to test the performance of our model on different settings of the calibration problem. The first synthetic problem has one functional calibration variable and one control variable. The second problem has an additional control variable, and the third problem has an additional global calibration variable. We also evaluate the performance of our model on a high dimensional synthetic problem (that is when $d^x$, the dimension of the domain of the calibration variables, is large), the results of which are presented in Appendix~D.} \BFcommTwo{We note that since in practical settings physical observations are noise contaminated, we add a noise drawn from $\mathcal{N}(0,0.05)$ to each generated $y^p$.}

\begin{table}[H]\footnotesize
\centering
\begin{tabular}{c| c| c | c|c}
Problem& $\mathcal{F}^s(\mathbf{x},\Btheta,\Bpsi)$ & $\mathcal{F}^p(\mathbf{x})$&  $\mathcal{F}^{\theta}(\mathbf{x})$&$\Bpsi$\\
\hline\hline
1&$\theta\exp(-0.05x^2)(\sin(x)^2+1)$&$\exp(-0.05x^2-0.05x)*(\sin(x)^2+1)$&$\exp(-0.05x)$&-\\
2&$0.4(x_1^2+x_2^2)\sin^2(0.7x_2)\frac {x_1+x_2}{\theta^2+1}$&$0.4(x_1^2+x_2^2)\sin^2(0.7x_2)$&$(x_1+x_2-1)^{0.5}$&-\\
3&$\theta+\psi x^2$&$2\sqrt{x}+2.5x^2$&$2\sqrt{x}$&2.5  \\
\hline
\end{tabular}
\caption{Calibration functions defined for the three synthetic problems}
\label{table:SynProbsFunctions}
\end{table}

\BFcomm{For the first synthetic problem, we locate $m=6$ control vectors, $\mathbf{x}^p$, at locations $\{0.5,1.5,2.5,3.5,$  $4.5,5.5\}$. Then for each $\mathbf{x}^p$, we randomly sample five functional calibration vectors from the interval $[0,2]$; therefore, we have a total of $n=30$ computational data points. Finally, we sample 12 random test control vectors, $\mathbf{x}^*$, from the line segment $[0,6]$ to form a  test dataset.}

For the second synthetic problem, we locate $m=16$ control vectors, $\mathbf{x}^p$, uniformly on the square $[0,3.5]\times[0,3.5]$. Then for each $\mathbf{x}^p$, we sample 10 functional calibration vectors randomly from the interval $[0,5]$; therefore, we have a total of $n=160$ computational data points. Finally, we sample  10 random test control vectors, $\mathbf{x}^*$, from the same square $[0,3.5]\times[0,3.5]$ to form a  test dataset.

For the third synthetic problem, we follow the setting used by~\cite{NBC}, where we choose $m=15$ control vectors for training at locations $\{0,0.05,0.10,0.15,0.20,0.25,0.30,$     $0.35,0.40,0.70,0.75,0.80,0.85,0.90,0.95\}$, and use five physical control vectors for testing at locations $\{0.45,0.50,0.55,0.60,0.65\}$. We sample 10 functional calibration vectors for each $\mathbf{x}^p$ from the square $[0,5]\times[0,5]$; therefore, we have a total of $n=150$ computational data points.

 The first real problem from a spot welding application was originally introduced by~\cite{bayarri2007framework}. This problem has three control variables and one calibration variable. The dataset associated with spot welding contains 12  and 35 data points sampled from the physical experiments and the computation model, respectively. The second real problem studied by~\cite{PFC} has one control variable and one calibration variable, and the associated dataset contains 11 and 150 data points sampled from the physical experiments and the computation model, respectively. This instance arises from a PVA-treated buckypaper fabrication process. 

For the real problems, we partition the sets of physical data points using four-fold cross validation to form training and test datasets. Therefore, for each iteration of cross validation for the spot welding dataset, we have eight physical data points in the training set and four physical data points in the test set. Similarly, for the PVA dataset we have eight to nine physical data points in the training set, and two to three data points in the test set in each iteration of cross validation. We note that the cross validation does not affect the size of the computational datasets, i.e., $n=35$ for spot welding and $n=150$ for the PVA dataset.

\subsection{Results}

We compare the results of our proposed methods with competing functional calibration methodologies. These include non-parametric functional calibration (NFC)~\citep{NFC}, parametric functional calibration (PFC) ~\citep{PFC}, and non-parametric Bayesian calibration  (NBC)~\citep{NBC}, which all require surrogate modeling for handling expensive computational models. When we use \apcomm{the generalized minimum spanning tree model on a calibration digraph to find a set of anchor points, we refer to our approach as BMNC. If we use the shortest path model on a calibration digraph, we call the approach BMNC-DAG.}

For each of the aforementioned problems, we choose the values of $\lambda$, $r$, and $M$ following the empirical approach explained in Section~\ref{sec.results} (See   Table~\ref{table:CalGraphParams}). Tables~\ref{table:RMSEs} \BFcomm{and~\ref{table:ComputationTimeTable}} compare the performance of BNMC and BNMC-DAG in terms of $\text{RMSE}_\theta$, $\text{RMSE}_y$, and \BFcomm{computation time} with the other competing methodologies. 

For the first synthetic problem, the second and third columns of Table~\ref{table:RMSEs} show that BNMC and BNMC-DAG both perform more accurately than the other methodologies in terms of $\text{RMSE}_{y}$, and they have the same order of accuracy in terms of $\text{RMSE}_{\theta}$. \BFcommTwo{Moreover, Table 4 shows that BNMC-DAG performs slightly faster than BNMC. However, because we only have $n=30$ computational and $m=6$ physical data points, this difference is not very large. To better compare the computational costs of the approach using the shortest path and the GMST models, we choose values of $n$s from the set $\{100,200,300,400, 500\}$ and run both BNMC and BNMC-DAG. As expected, the shortest path model in BNMC-DAG finds the anchor points for all the values of $n$ in less than a second, whereas GMST requires $3.6, 21.2, 65.3,144.7, \text{ and }323.5$ seconds for $n \in \{ 100,200,300,400,500\}$, respectively.}

For the second synthetic problem, because the dimension of $\mathbf{x}^p$ is greater than one, we cannot apply BNMC-DAG. However, the fourth and fifth columns of Table~\ref{table:RMSEs} show that BNMC outperforms the other methodologies in terms of $\text{RMSE}_{y}$ and has the second best accuracy in terms of $\text{RMSE}_{\theta}$.

For the third synthetic problem, we only compare the results of NBC and BNMC, since the codes for NFC and PFC are written only for univariate calibration problems. The sixth and the seventh columns of Table~\ref{table:RMSEs} show that BNMC outperforms NBC both in terms of $\text{RMSE}_y$ and $\text{RMSE}_\theta$. We note that the reported $\text{RMSE}_y$ for NBC in~\citep{NBC} under the cheap computational code assumption is 0.0538, which is a better accuracy compared to that of  BNMC; however, here BNMC is superior when NBC uses surrogate modeling. 

Since the true values of the calibration parameters are unknown for the real problems, we compare the results only in terms of $\text{RMSE}_y$. The eighth column (PVA) of Table~\ref{table:RMSEs} shows that BNMC, BNMC-DAG, and NBC have the same order of accuracy and perform better than NFC and PFC. Finally, we observe in the last column of Table~\ref{table:RMSEs} that for the Spot Welding problem BNMC outperforms the other competing methodologies with a large margin. We attribute this performance to the capability of BNMC in handling expensive computational models with a small number of computational data points.

\begin{table}[H]\footnotesize
\centering
\begin{tabular}{c| c| c | c |c|c}
Parameter& $1^{\text{st}}$ synthetic problem & $2^{\text{nd}}$ synthetic problem&$3^{\text{rd}}$ synthetic problem& PVA& Spot Welding\\
\hline\hline
$\lambda$&0.5&0.5&0.5&0.5&0.5\\
$r$&1.5&0.9&0.16&0.5&4\\
$M$&$10^5$&$10^5$&$10^5$&$10^5$&$10^5$\\
\hline
\end{tabular}
\caption{The calibration graph parameters for the five calibration problems.}
\label{table:CalGraphParams}
\end{table} 

\begin{table}[H]\footnotesize
\centering
\begin{tabular}{c| c | c| c |c |c |c|c|c}
Methodology&\multicolumn{2}{c|}{$1^{\text{st}}$ synthetic problem} &\multicolumn{2}{c|}{ $2^{\text{nd}}$ synthetic problem}&\multicolumn{2}{c|}{$3^{\text{rd}}$ synthetic problem} & PVA& Spot Welding\\
\hline\hline
{} & $\text{RMSE}_y$&$\text{RMSE}_\theta$&$\text{RMSE}_y$&$\text{RMSE}_\theta$&$\text{RMSE}_y$&$\text{RMSE}_\theta$&$\text{RMSE}_y$&$\text{RMSE}_y$\\
\hline
NFC		 &0.184&0.254&0.143&0.202&-    &-    &0.379&0.683\\
PFC	 	 &0.226&0.244&0.296&0.390&-    &-    &0.450&1.115\\
NBC	 	 &0.162&0.356&0.132&0.627&0.172&0.426&0.281&0.516\\
BNMC	 &0.098&0.271&0.076&0.356&0.063&0.354&0.296&0.409\\
BNMC-DAG&0.098&0.265  & -& -&- &- &0.288&-    \\

 \hline
\end{tabular}
\caption{$\text{RMSE}_\theta$ and $\text{RMSE}_y$ of different methodologies for the five calibration problems.}
\label{table:RMSEs}
\end{table}

\begin{table}[H]\footnotesize
\centering
\begin{tabular}{c| c| c | c |c|c}
Methodology& $1^{\text{st}}$ synthetic problem & $2^{\text{nd}}$ synthetic problem&$3^{\text{rd}}$ synthetic problem& PVA& Spot Welding\\
\hline\hline
NFC	    & 2.88&  8.02 &	-     &  3.29&0.77  \\
PFC	    &18.04&253.08 &	-     &176.95&19.06 \\
NBC     &45.26&307.31 &	230.19&180.92&68.04 \\
BNMC	&125.15&169.23&	147.01&159.84&117.53\\
BNMC-DAG&123.91&	- &	-	  &144.54&-     \\
\hline
\end{tabular}
\caption{The computation times (in seconds) for the five calibration problems.}
\label{table:ComputationTimeTable}
\end{table} 

\begin{figure}[H]
\begin{center}
\begin{subfigure}{0.45\textwidth}
		\includegraphics[height=4.5cm,width=7cm]{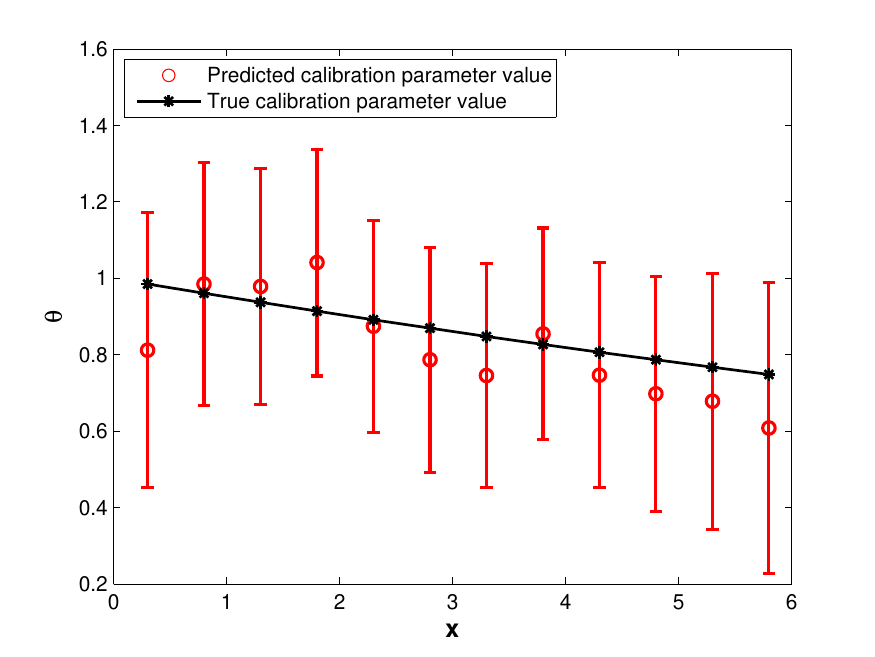}
		\caption {\tiny $1^{\text{st}}$ synthetic problem calibration plot}
\label{fig:Small1DSynProb_ThetaCI}
	\end{subfigure}
	\begin{subfigure}{0.45\textwidth}
		\includegraphics[height=4.5cm,width=7cm]{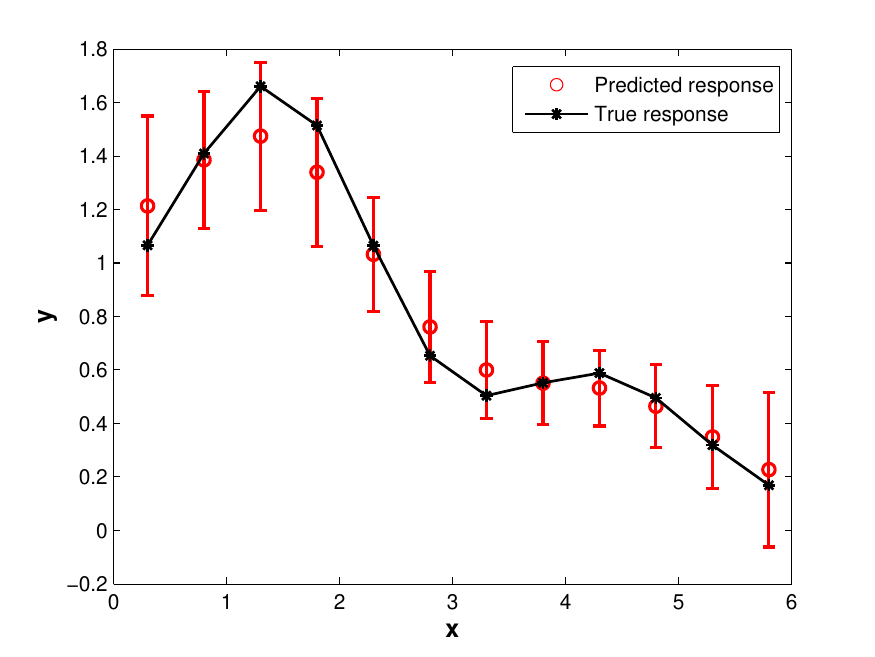}
		\caption {\tiny $1^{\text{st}}$ synthetic problem response plot}
\label{fig:Small1DSynProb_YCI}
\end{subfigure}

	\begin{subfigure}{0.45\textwidth}
		\includegraphics[height=4.5cm,width=7cm]{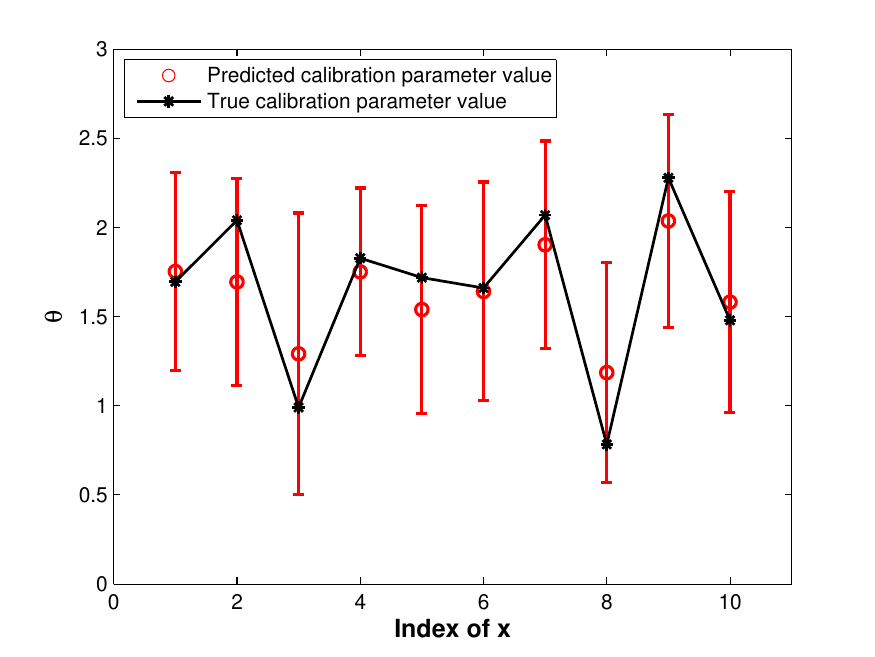}
		\caption {\tiny $2^{\text{nd}}$ synthetic problem calibration plot}
\label{fig:Syndata1_ThetaCI}
	\end{subfigure}
	\begin{subfigure}{0.45\textwidth}
		\includegraphics[height=4.5cm,width=7cm]{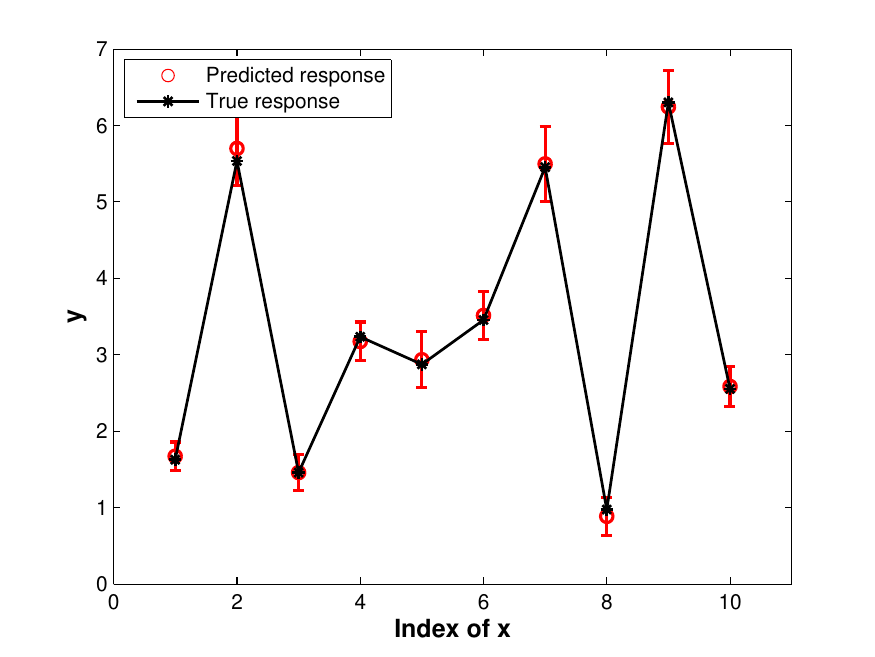}
		\caption {\tiny $2^{\text{nd}}$ synthetic problem response plot}
\label{fig:Syndata1_YCI}
	\end{subfigure}
	
	\begin{subfigure}{0.45\textwidth}
		\includegraphics[height=4.5cm,width=7cm]{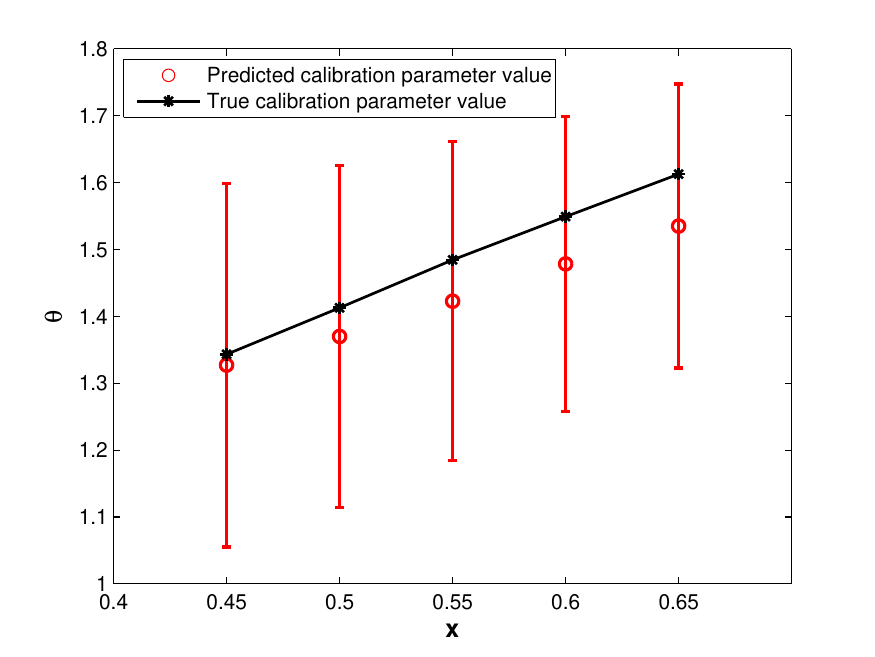}
		\caption {\tiny $3^{\text{rd}}$ synthetic problem calibration plot}
\label{fig:Syndata2_ThetaCI}
	\end{subfigure}
	\begin{subfigure}{0.45\textwidth}
		\includegraphics[height=4.5cm,width=7cm]{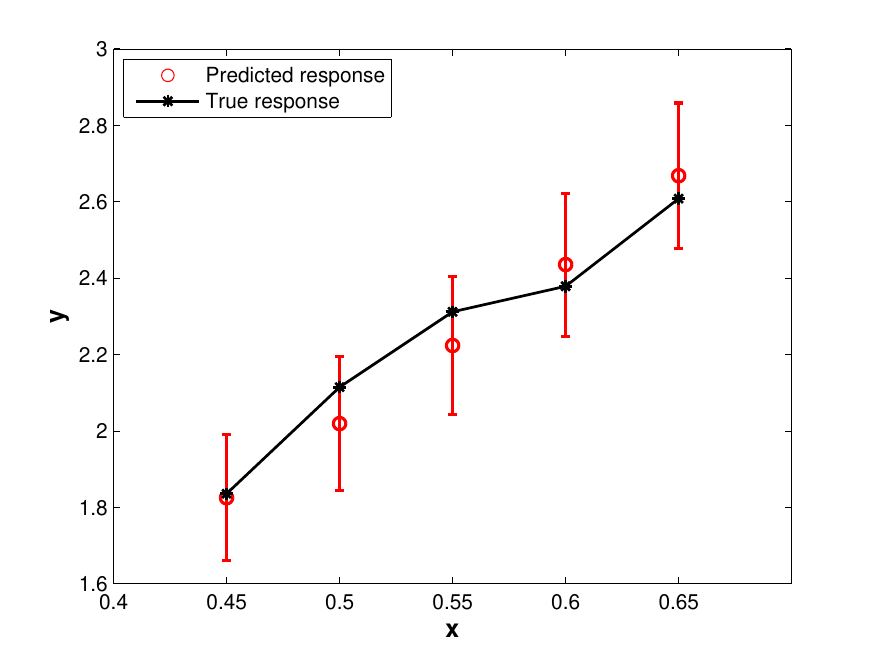}
		\caption {\tiny $3^{\text{rd}}$ synthetic problem response plot}
\label{fig:Syndata2_YCI}
\end{subfigure}
\end{center}
\caption{The $95\%$ confidence interval predictions for functional calibration parameter values and responses for the test datasets of the synthetic problems.}
\label{fig:Syndata_CI}
\end{figure}

Figure~\ref{fig:Syndata_CI} shows the $95\%$ confidence interval predictions for the responses and the functional calibration parameter for the test datasets of the  synthetic problems. Note that since $\mathbf{x}^p\in \rl^2$ for the second synthetic problem, we plot the predicted values against their indices in Figures~\ref{fig:Syndata1_ThetaCI} and~\ref{fig:Syndata1_YCI}, and connect the data points to each other for better visualization. \BFcommTwo{Moreover, although we added white noise to response variables to mimic real world processes, we show the denoised responses for clearer illustration.} 

As noted in Section~\ref{sec.BayesianModel}, due to the limited number of samples we collect from the computational model, we cannot accurately recover $\mathcal{F}^{\theta}$, but the way we train the hyper-parameters of the GP aims to compensate for this limitation. We can observe this in Figure~\ref{fig:Syndata_CI}, where the predictions of the response values have better accuracy and tighter confidence intervals compared to those of functional calibration parameter values.

\begin{figure}[H]
\begin{center}
    \begin{subfigure}{0.45\textwidth}
		\includegraphics[height=4.5cm, width=7cm]{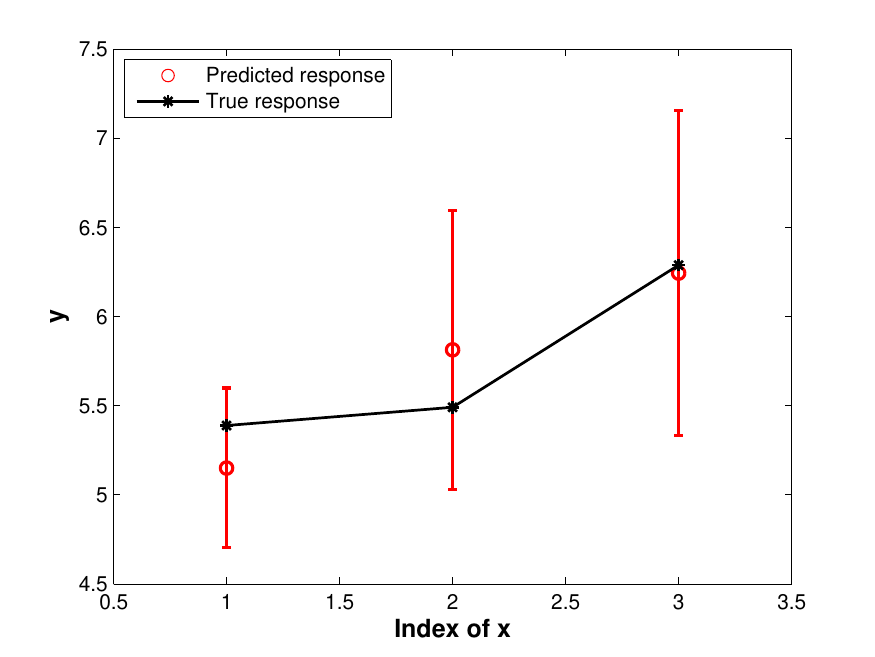}
        \caption{\tiny  Spot welding response plot}
        \label{fig:Weld_CI}
	\end{subfigure}
		\begin{subfigure}{0.45\textwidth}
		\includegraphics[height=4.5cm, width=7cm]{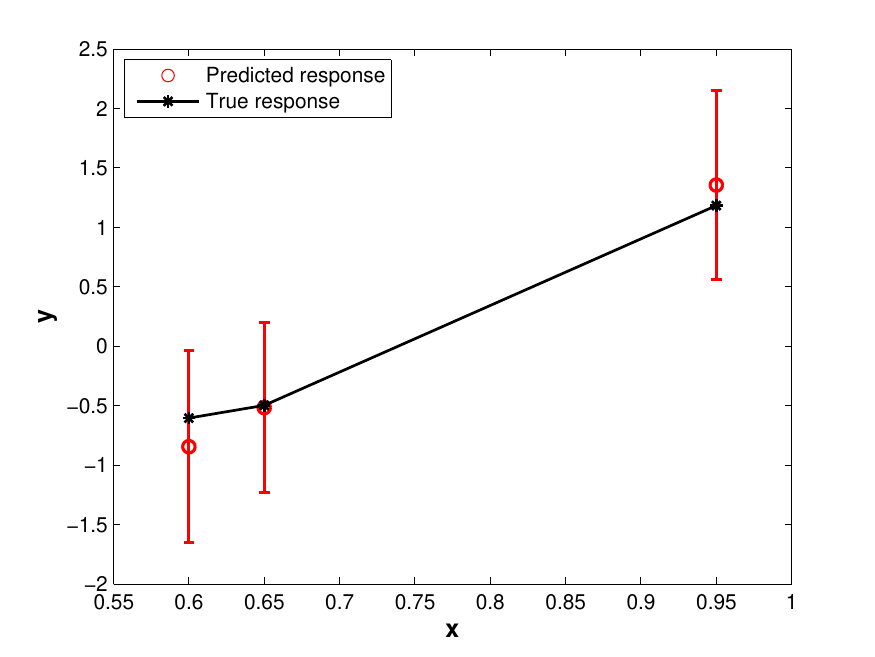}
        \caption{\tiny PVA response plot}
        \label{fig:PVA_CI}
	\end{subfigure}
\end{center}
\caption{The $95\%$ confidence interval predictions for one of the test datasets created in the cross validation process for each of the spot welding and the PVA problems.}
\label{fig:PVAandSpotWelding}
\end{figure}

For illustration, Figure~\ref{fig:PVAandSpotWelding} shows the $95\%$ confidence interval predictions for one of the test datasets created in the cross validation process for each of the spot welding and the PVA problems. Since we do not know the true functional calibration parameter values, we cannot provide a similar plot for the calibration predictions. Similar to Figures~\ref{fig:Syndata1_ThetaCI} and~\ref{fig:Syndata1_YCI}, we plot the predicted values against their indices in Figure~\ref{fig:Weld_CI} for better visualization. 

%As noted in Section~\ref{sec.BayesianModel}, due to the limited number of samples we collect from the computational model, we cannot accurately recover $\mathcal{F}^{\theta}$, but the way we train the hyper-parameters of the GP aims to compensate for this limitation. We can observe this in Figure~\ref{fig:Syndata_CI}, where the predictions of the response values have better accuracy and tighter confidence intervals compared to those of functional calibration parameter values.

%For illustration, Figure~\ref{fig:PVAandSpotWelding} shows the $95\%$ confidence interval predictions for one of the test datasets created in the cross validation process for each of the spot welding and the PVA problems. Since we do not know the true functional calibration parameter values, we cannot provide a similar plot for the calibration predictions. Similar to Figures~\ref{fig:Syndata1_ThetaCI} and~\ref{fig:Syndata1_YCI}, we plot the predicted values against their indices in Figure~\ref{fig:Weld_CI} for better visualization. 

\BFcommTwo{We refer the reader to Appendix~E for an analysis of the residuals to validate the assumptions made in our proposed approach, especially those made in equations~\eqref{ComPhyRel} and~\eqref{NonVecModel}.}

\section{Concluding remarks}\label{sec.concl}
We proposed a Bayesian non-isometric matching calibration model for expensive computational models. A limited budget to evaluate computational models led to the use of a GP, which was trained \emph{during} the calibration procedure. We used Bayesian statistics to simultaneously train the hyper-parameters of the GP's covariance function and make inferences on the calibration parameters associated with the physical data points.
To construct informative prior distributions for our new approach, we used a geometric interpretation of calibration based on non-isometric curve to surface matching. This point of view enabled us to develop graph-theoretic approaches to address the problem of finding a set of anchor points  used in constructing informative prior distributions. For the special case of a single control and calibration variable, we introduced a shortest path model on a directed acyclic calibration graph to tackle the problem of finding anchor points, while for the general case, we introduced the generalized minimum spanning tree model. Our numerical experiments conducted on four benchmark calibration problems showed that our approach outperformed the existing calibration models under the assumption of expensive computational models.

The framework developed in this paper could be extended in several ways. We only considered a single computational data point to construct a prior distribution for each calibration parameter; however, information of multiple computational data points could be taken into account. An implementation of this idea, of course, requires developing new combinatorial optimization techniques capable of choosing an appropriate number of computational data points. Another interesting research path would be to consider data uncertainty formally in the calibration graph model instead of using a deterministic calibration graph model, and then using Bayesian inference to deal with the uncertainty in the data. \BFcommThree{Moreover, one may consider including an independent discrepancy function in equation~\eqref{NonVecModel}, as noted in Section~\ref{sec.BayesianModel}.} \apcomm{Finally, the proposed approach would potentially benefit from using cross-validation for the selection of the tuning parameters such as $\lambda$, $r$, and $M$.}

\section*{Acknowledgment}
Austin Buchanan is supported by the National Science Foundation under Grant No.\,1662757. This work was completed utilizing the High Performance Computing Center facilities of Oklahoma State University at Stillwater. The authors also acknowledge \alert{Chuck Zhang} from Georgia Institute of Technology for providing the authors with the PVA-treated buckypaper fabrication data.

\onehalfspacing
\bibliographystyle{chicago}
\bibliography{bibfile}
\newpage
\doublespacing
\begin{appendices}

\section{Full and conditional posterior distributions}\label{PosteriorandFullCondPosteriors}
First, we explain the prior distributions for parameters in~\eqref{posterior_concise}. We have
\begin{eqnarray*} 
\begin{aligned}
%& \BTheta^{p}_k\suchthat\BTheta^{a}_k,{\nu}_k^2\ \sim \mathcal{N}(\BTheta^{a}_k,\nu^2_k \BI_m),&\forall k\in [d^{\theta}],\\
%&\Bpsi^p\suchthat\Bpsi^a,\Btau^2\sim \mathcal{N}(\Bpsi^a, \diag(\Btau^2)),\\
& \nu^2_k \sim \frac{1}{\nu_k^2},&\forall k\in \{1,\ldots,d^{\theta}\},\\
& \tau^2_h \sim \text{Inv-Gamma} (\alpha_{\tau},\beta_{\tau}),&\forall h\in\{1,\ldots,d^{\psi}\},\\
& \ell_j\sim \text{Log-Gamma}(\alpha_{\ell},\beta_{\ell}), &\forall j\in \{1,\ldots,d^x+d^{\theta}+d^{\psi}\},\\
& \gamma\sim \text{Log-Uniform},\\
& \sigma^2\sim \text{Log-Uniform}.
\end{aligned}
\end{eqnarray*} 

For each $\nu^2_k$ we use a flat Jeffreys prior~\citep{jeffreys1946invariant}, which is an inverse gamma distribution with zero value for both the shape and the scale parameter. For each $\tau^2_h$ we choose a weak inverse gamma distribution, i.e., an inverse gamma with large variance,  with $\alpha_{\tau}=2.1$ and $\beta_{\tau}=10$ as its parameters. Note that both of these prior distributions are conjugate for their associated parameters in the posterior distribution. Moreover, as recommended by~\cite{gelman2014bayesian}, to improve the identifiability of the model we use the prior distributions on the logarithmic scale for parameters of the GP part of the model. Therefore, for $\sigma^2$ and $\gamma$ we use a flat log-uniform distribution and for each $\ell_j$ we use a log-gamma distribution with the parameters $\alpha_{\ell}=\beta_{\ell}=2$.

Using the prior distributions explained above the posterior~\eqref{posterior_concise} can be written as
\begin{eqnarray}
\begin{aligned}
& \pi(\BTheta^p,\Bpsi^p,\Bnu^2, \Btau^2,\BEll,\gamma,\sigma^2\suchthat\mathbf{y}^p,\mathbf{X}^p,\BTheta^a,\Bpsi^a) \\
&\propto|\BSigma+\sigma\BI_m|^{-0.5}\exp\left\{\frac{-1}{2}{\mathbf{y}^p}^{\top} (\BSigma+\sigma\BI_m)^{-1}\mathbf{y}^p\right\}\\
& \times \prod_{k} (\nu^2_k)^{-m/2-1} \exp\left\{\frac{-1}{2\nu^2_k}(\BTheta^{p}_k-\BTheta^{a}_k)^{\top}(\BTheta^{p}_k-\BTheta^{a}_k)\right\}\\
&\times \prod_{h} (\tau^2_h)^{-1/2-\alpha_{\tau}-1} \exp\left\{\frac{-1}{2\tau^2_h}(\psi_h^p-\psi^a_h)^2\right\}\exp\left\{\frac{-\beta_{\tau}}{\tau_h^2}\right\}\\
&\times \prod_{j}\frac{1}{\ell_j}\log(\ell_j)^{\alpha_{\ell}-1}\exp\left\{-\beta_{\ell}\log(\ell_j)\right\}\\
&\times \frac{1}{\gamma}\times \frac{1}{\sigma^2}.\label{Posterior4}
\end{aligned}
\end{eqnarray}

We use Gibbs sampling~\citep{gelfand1990illustration} to sequentially sample from the full conditional posterior distributions. Here we present the full conditional distribution for each of the parameters in~\eqref{Posterior4}:

\begin{eqnarray*}
\begin{aligned}
&\pi(\Theta_{ki}^p\suchthat \cdot)
    \propto|\BSigma+\sigma\BI_m|^{-0.5}\\
    &\;\;\;\;\;\;\;\;\;\;\;\;\;\;\;\times \exp\left\{\frac{-1}{2}\Big({\mathbf{y}^p}^{\top} (\BSigma+\sigma\BI_m)^{-1}\mathbf{y}^p+\frac{1}{\nu^2_k}(\Theta^{p}_{ki}-\Theta^{a}_{ki})^2\Big)\right\}&\forall k\in \{1,\ldots,d^{\theta}\},\forall i \in\{1,2,\ldots,m\}\\
&\pi(\psi^p_h\suchthat \cdot)\propto|\BSigma+\sigma\BI_m|^{-0.5}\\
&\;\;\;\;\;\;\;\;\;\;\;\;\;\;\;\times\exp\left\{\frac{-1}{2}\Big({\mathbf{y}^p}^{\top} (\BSigma+\sigma\BI_m)^{-1}\mathbf{y}^p+\frac{1}{\tau^2_k}(\psi_h^p-\psi^a_h)^2\Big)\right\}&\forall h\in\{1,\ldots,d^{\psi}\}\\
&\pi(\nu^2_k\suchthat \cdot)\propto (\nu^2_k)^{-m/2-1} \exp\left\{\frac{-1}{2\nu^2_k}(\BTheta^{p}_k-\BTheta^{a}_k)^{\top}(\BTheta^{p}_k-\BTheta^{a}_k)\right\}&\forall k\in \{1,\ldots,d^{\theta}\}\\
&\pi(\tau^2_h\suchthat \cdot)\propto (\tau^2_h)^{-1/2-\alpha_{\tau}-1} \exp\left\{\frac{-1}{2\tau^2_h}\Big((\psi_h^p-\psi^a_h)^2+2\beta_\tau\Big)\right\}&\forall h\in\{1,\ldots,d^{\psi}\}\\
&\pi(\ell_j\suchthat \cdot)\propto \frac{\log(\ell_j)^{\alpha_\ell-1}|\BSigma+\sigma\BI_m|^{-0.5}}{\ell_j}\\ 
&\;\;\;\;\;\;\;\;\;\;\;\;\;\;\;\times\exp\left\{\frac{-1}{2}\Big({\mathbf{y}^p}^{\top} (\BSigma+\sigma\BI_m)^{-1}\mathbf{y}^p+2\beta_\ell\log(\ell_j)\Big)\right\} &\forall j\in \{1,\ldots,d^x+d^{\theta}+d^{\psi}\}\\
&\pi(\gamma\suchthat \cdot)\propto \frac{|\BSigma+\sigma\BI_m|^{-0.5}}{\gamma}\exp\left\{\frac{-1}{2}{\mathbf{y}^p}^{\top} (\BSigma+\sigma\BI_m)^{-1}\mathbf{y}^p\right\}\\
&\pi(\sigma^2\suchthat \cdot)\propto \frac{|\BSigma+\sigma\BI_m|^{-0.5}}{\sigma^2}\exp\left\{\frac{-1}{2}{\mathbf{y}^p}^{\top} (\BSigma+\sigma\BI_m)^{-1}\mathbf{y}^p\right\},
\end{aligned}
\end{eqnarray*}
where the notation $(\suchthat\cdot)$ denotes the conditioning over every other parameters of posterior~\eqref{Posterior4}.

We note that $\nu_k^2\suchthat\cdot$ and $\tau_k^2\suchthat\cdot$ have inverse gamma distributions with parameters $\Big(\frac{m}{2},\frac{(\BTheta^{p}_k-\BTheta^{a}_k)^{\top}(\BTheta^{p}_k-\BTheta^{a}_k)}{2}\Big)$ and $\Big(\frac{1}{2}+\alpha_\tau,\frac{(\psi_h^p-\psi^a_h)^2+2\beta_\tau}{2}\Big)$, respectively. However, the rest of the conditional distributions do not have closed form distributions; therefore, we  take Metropolis-Hastings~\citep{metropolis1953equation} steps to sample from these distributions during the sampling process.

\section{Proof of Lemma~\ref{lemma_dag} and Proposition~\ref{Prop_ThetaStarPred}}\label{Proof_Prop_ThetaStarPred}

\begin{proof}[\textbf{Proof of Lemma~\ref{lemma_dag}}]
It suffices to show  that if $(u,v)$ is an edge, then $u < v$, which is trivially true when $u=0$ or $v = n+1$. For distinct vertices $u,v \in V^0$, note that $u < v$ if and only if $\mathbf{x}^s_u < \mathbf{x}^s_v$ as we have assumed the points in $S$ to be strictly ordered. Suppose,  $u\in C_i$ and $v\in C_{i+1}$ for some $i \in \{1,2,\ldots,m-1\}$. Hence, $\mathbf{x}^p_i < \mathbf{x}^p_{i+1}$, and \comm{from equation}~\eqref{eq:cluster_rule} we can conclude that the distinct points $\mathbf{x}^s_u$ and  $\mathbf{x}^s_v$  satisfy $\mathbf{x}^s_u \le \frac{1}{2}(\mathbf{x}^p_i+ \mathbf{x}^p_{i+1})\le  \mathbf{x}^s_v.$
\end{proof}

\begin{proof}[\textbf{Proof of Proposition~\ref{Prop_ThetaStarPred}}]
To obtain the distribution of $\BTheta^{p}_k(t)$, we note that the assumptions $[\epsilon^\theta_1,\ldots,\epsilon^\theta_m]^{\top}\sim \mathcal{N}(0,\sigma^\theta_k\BI_m)$ and $\mathcal{F}^\theta_k \sim \mathcal{GP}(0,\mathcal{K(\cdot,\cdot)})$ imply that
\begin{eqnarray}
\BTheta^{p}_k(t)\sim\mathcal{N}(0,\BSigma_{\mathbf{X}^p\mathbf{X}^p}+\sigma^\theta_k\BI_m),\label{MargnialLikeTheta}
\end{eqnarray}
where we denote the covariance between columns of matrices  $\mathbf{Z}$ and $\mathbf{Z}'$ by $\BSigma_{\mathbf{Z}\mathbf{Z}'}$. We add the variance of error, $\sigma_k^{\theta}$, to preserve the smoothness of $\mathcal{F}^\theta_k$; otherwise, our approach would obtain $\mathcal{F}^\theta_k$ as the interpolation of the elements of $\BTheta^p_k(t)$.

Further, by the GP assumption on $\mathcal{F}^\theta_k$, we obtain the joint distribution of $\BTheta^{p}_k(t)$ and the prediction of $\theta^*_{k}$ for the $t^{th}$ draw, which we denote by $\theta_k^*(t)$, as
\begin{equation}
\begin{bmatrix}
\BTheta^{p}_k(t) \\
\theta_{k}^*(t)
\end{bmatrix}
\sim \mathcal{N}
\left(0,
\begin{bmatrix}
\BSigma_{\mathbf{X}^p\mathbf{X}^p}+\sigma^{\theta}_k\BI_m & \BSigma_{\mathbf{X}^p\mathbf{x}^*}\\
\BSigma_{\mathbf{x}^*\mathbf{X}^p}& \BSigma_{\mathbf{x}^*\mathbf{x}^*}
\end{bmatrix}
\right).\label{JointLikTheta}
\end{equation}
By conditioning on  $\BTheta^p_k(t)$ in~\eqref{JointLikTheta}, the point prediction of $\theta_{k}^*(t)$ is obtained as
\begin{eqnarray*}
\theta_{k}^*(t)=\BSigma_{\mathbf{x}^*\mathbf{X}^p}(\BSigma_{\mathbf{X}^p\mathbf{X}^p}+\sigma^{\theta}_k\BI_m)^{-1}\BTheta^{p}_k(t).
\end{eqnarray*}

\end{proof}

We note that for each prediction in~\eqref{PredTheta}, the hyper-parameters of the covariance function used in~\eqref{MargnialLikeTheta} should be tuned, which can be achieved by maximizing the logarithm of likelihood corresponding to~\eqref{MargnialLikeTheta}~\citep{rasmussen2004gaussian}:
\begin{eqnarray*}
\log(\pi({\Theta}^{p}_k(t)))=-\frac{n}{2}\log(2\pi)-\frac{1}{2}\log|\BSigma_{\mathbf{X}^p\mathbf{X}^p}+\sigma^{\theta}_k\BI_m|-\frac{1}{2}\BTheta^{p^{\top}}_k(t)(\BSigma_{\mathbf{X}^p\mathbf{X}^p}+\sigma^{\theta}_k\BI_m)^{-1}\BTheta^{p}_k(t).
\end{eqnarray*}

\section{Integer programming formulations}\label{app.formuls}

The GMST problem can also be viewed as a special case of the \emph{generalized minimum spanning arborescence} (GMSA) problem, which is defined on a directed graph with its vertex set partitioned into clusters. Here we seek an arborescence (i.e., directed out-tree) of minimum weight, rooted  at some vertex in a specified cluster that contains exactly one vertex per cluster. We can transform the GMST problem on graph $G=(V,E)$ to the GMSA problem by replacing each undirected edge $\{i,j\} \in E$  with directed anti-parallel arcs $(i,j)$ and $(j,i)$. Then, each arc is assigned the same weight as the corresponding undirected edge, and we arbitrarily choose one of the clusters to contain the root. 

 For the remainder of this discussion, we  use  the  directed graph  $\overset{\leftrightarrow}{G}=(V,A)$ corresponding to the calibration graph $G=(V,E)$, where $A \coloneqq \bigcup\limits_{\{u,v\} \in E} \{ (u,v), (v,u)\}$. The edge weight of each $e = \{u,v\} \in E$  is duplicated as arc-weights $w_{uv} = w_{vu} \coloneqq w_e$. Recall that the vertex set $V$ is partitioned into clusters, say $C_1,\ldots,C_m$.   We require the arborescence to be rooted at some vertex in $C_1$.  We use binary decision vectors $q \in \{0,1\}^{|A|}$ and $b \in \{0,1\}^{|V|}$ to denote the incidence vectors of the arcs and vertices included in the arborescence, respectively. 
\begin{subequations}
\begin{align}
\textbf{(DCSUB)} \quad \min \sum_{(u,v)\in A} w_{uv} q_{uv}&  \\
\text{subject to: }\quad 
\sum_{v \in C_i} b_{v}&=1 \quad \forall i \in \{1,2,\ldots,m\}\label{DCSUB1}\\
q_{uv} + q_{vu} &\leq 1 \quad \forall (u,v), (v,u) \in A \label{DCSUB2}\\
\sum_{(u,v)\in A}q_{uv} &=m-1\label{DCSUB3}\\
\sum_{(u,v)\in A : u,v \in Q} q_{uv} &\leq \sum_{v\in Q}b_v -1 \quad \forall Q\subset V \suchthat Q \supset C_i \text{ for some } i\in \{1,2,\ldots,m\}\label{DCSUB4}\\
\sum_{u : (u,v) \in A}q_{uv} &= b_v \quad \forall v\in V\backslash C_1\label{DCSUB5}\\
 b &\in \{0,1\}^{|V|},q \in \{0,1\}^{|A|}.\label{DCSUB6}
\end{align}\label{form.DCSUB}
\end{subequations}
Constraints~\eqref{DCSUB1} enforce that the model  chooses exactly one vertex from each cluster, constraints~\eqref{DCSUB3} ensure that exactly $m-1$ arcs from $A$ are selected, and constraints~\eqref{DCSUB2} ensure that these correspond to $m-1$ distinct edges in $E$. Cluster subpacking constraints~\eqref{DCSUB4} prevent solutions that contain cycles and were shown by 
\cite{feremans2002comparative}  to dominate the  subtour elimination constraints used by \cite{myung1995generalized}: \[\sum_{(v,w)\in A : v,w \in S} q_{vw} \leq \sum_{v\in S\setminus\{u\}}b_v  \quad \forall u \in S\subset V,\ 2 \le |S| \le |V|-1.\] Finally, constraints~\eqref{DCSUB5} ensure that every non-root vertex selected by the solution has exactly one incoming edge and every vertex outside $C_1$ that is not selected will have no incoming arcs.  Combined with the requirement that we choose exactly $m$ vertices and $m-1$ arcs without creating cycles, this ensures that we obtain an arborescence rooted at some vertex inside $C_1$. \comm{As discussed in Section~\ref{ssec.GMSTIP}, a computationally effective approach for solving the GMST problem using this formulation requires delayed constraint generation.} This approach starts by relaxing formulation~\eqref{form.DCSUB} by omitting constraints~\eqref{DCSUB4}, \comm{and adding them ``on the fly'' during the progress of the branch-and-bound algorithm}.

Next we present the multi-commodity flow (MCF) formulation for  the GMSA problem that avoids using exponentially many constraints, but uses an additional set of variables~\citep{myung1995generalized}. The MCF formulation treats every vertex $v \in C_1$ to have supply $b_v$ for each commodity $i \in \{2,\ldots,m\}$ corresponding to the remaining clusters; it treats every  $v \in C_i$ to have a demand of $b_v$ for commodity $i$. Suppose $b_v = 1$ for some $v \in C_i$, then a path must be traced from the root selected in $C_1$ to deliver commodity $i$. We use the additional set of commodity-flow variables $f_{uv}^i$ to denote the amount of commodity $i \in  \{2,3,\ldots,m\}$ flowing on arc $(u,v) \in A$. 
\begin{subequations}
\begin{align}
\textbf{(MCF)}\quad \min\sum_{(u,v)\in A} w_{uv}q_{uv} &  \\
 \text{subject to:} \quad 
\eqref{DCSUB1}, \eqref{DCSUB2}, \eqref{DCSUB3}, \eqref{DCSUB6} &\nonumber \\
\sum_{w : (v,w) \in A} f^i_{vw}-\sum_{u : (u,v) \in A}  f^i_{uv}& = \left\{ \begin{array}{rl}
    b_v, & \quad \forall v\in C_1 \\-b_v, &\quad  \forall v\in C_i\\0, &\quad \forall v \notin C_1 \cup C_i
\end{array}\right\} \quad\quad \forall i \in  \{2,\ldots,m\} \label{MCF4}\\
0 \le f^i_{uv}& \leq q_{uv} \quad\quad \forall (u,v) \in A, i \in  \{2,\ldots,m\}. \label{MCF5}
\end{align}\label{form.MCF}
\end{subequations}
Constraints~\eqref{MCF4} are flow-balance constraints for each commodity, and constraints~\eqref{MCF5} prevent flows on the edges that are not selected. 

%\alert{Babak-- can you check in the original formulation and make sure that constraints~\eqref{DCSUB5} are not needed here? I think constraints~\eqref{MCF4} take care of the requirement; but let's double-check.} \BFcomm{I think 18f is needed. 18f is equivalent to constraint 9 in Fereman's paper (page 10)}

Formulations~\eqref{form.DCSUB} and~\eqref{form.MCF} are both equally good in terms of the tightness of the LP relaxations as the projection of the LP relaxation of the latter onto the $(b,q)$-space is the same as the LP relaxation of  the former~\citep{feremans2002comparative}.

\section{A high dimensional case study}\label{HighDimProbSec}
In this section, we evaluate the performance of the competing calibration methodologies on a $10$ dimensional problem, that is when the dimension of the domain of the control variables, $d^x=10$. For this problem, we assume that the computational model has the form
\[
\mathcal{F}^s(\mathbf{x},\Btheta,\Bpsi)=\mathcal{G}(\mathbf{x})+\Btheta,
\]
where $\mathcal{G}:[0,1]^{10}\longrightarrow \rl$ is a Gaussian process with a zero mean  and covariance function~\eqref{SqrExpKernel} with parameters $\gamma=1$ and $\ell^T=[11.1,6.2,4,2.7,2,1.5,1.2,1,0.8,0.6]$. To build $\mathcal{G}$, we first generate 5,000 samples from the cube $[0,1]^{10}$ and form the matrix $\mathbf{X}$ whose size is $5000\times 10$. We then generate 1000 samples from $\mathcal{N}(0,1)$ and form the vector $\mathbf{z}$. With the fixed matrix $\mathbf{X}$ and vector $\mathbf{z}$, we can explicitly define function $\mathcal{G}$ as (see~\cite{rasmussen2004gaussian} for detail)
\[
\mathcal{G}(\mathbf{x})=\mathbf{k}_{\mathbf{x}\mathbf{X}}(\mathbf{K}_{\mathbf{X}\mathbf{X}})^{-1}\mathbf{L}\mathbf{z},
\]
where $\mathbf{k}_{\mathbf{x}\mathbf{X}}$ is the covariance vector between $\mathbf{x}$ and $\mathbf{X}$, $\mathbf{K}_{\mathbf{X}\mathbf{X}}$ is the covariance matrix between $\mathbf{X}$ and $\mathbf{X}$, and $\mathbf{L}$ is the lower Cholesky decomposition of $\mathbf{K}_{\mathbf{X}\mathbf{X}}$.

We further define the physical model as $\mathcal{F}^p(\mathbf{x})=\mathcal{G}(\mathbf{x})+\sqrt{\mathbf{a}^T\mathbf{x}}$, where $\mathbf{a}^T=[0.1,0.3,0.5,0.7,0.9,$ $1.1,1.3,1.5,1.7,1.9]$ is a vector of coefficients. By definition of $\mathcal{F}^s(\mathbf{x},\Btheta,\Bpsi)$ and  $\mathcal{F}^p(\mathbf{x})$, we subsequently get $\mathcal{F}^{\Btheta}(\mathbf{x})=\sqrt{\mathbf{a}^T\mathbf{x}}$.

To form the computational and physical datasets, we locate $m=30$ control vectors, $\mathbf{x}^p$, uniformly on the square $[0,1]^{10}$. Then for each $\mathbf{x}^p$, we sample 10 functional calibration vectors randomly from the interval $[0,3]$; therefore, we have a total of $n=300$ computational data points. Finally, we sample  100 random test control vectors, $\mathbf{x}^*$, from the same square $[0,1]^{10}$ to form a  test dataset. Table~\ref{table:HighDimCaseTBL} shows the performace of BNMC, NBC, NFC, and PFC in terms of RMSE and computation time.

\begin{table}[H]\footnotesize
\centering
\begin{tabular}{c| c| c | c}
Methodology& $\text{RMSE}_y$& $\text{RMSE}_\theta$&Computation time (in seconds)\\
\hline\hline
NFC&0.89&0.21&60.19\\
PFC&0.89&0.24&1595.03\\
NBC&0.94&1.62&663.57\\
BNMC&0.90&0.18&1256.5\\
\hline
\end{tabular}
\caption{$\text{RMSE}_y$, $\text{RMSE}_\theta$, and computation time of different methodologies for the 10 dimensional problem}
\label{table:HighDimCaseTBL}
\end{table} 

Due to the fact that the input space is large, and we are using a small number of samples, none of the competing methodologies have an advantage over the others in terms of $\text{RMSE}_y$; however, BNMC outperforms the other methodologies in terms of $\text{RMSE}_\theta$. \apcommtwo{However, when we consider both $\text{RMSE}_y$ and time, NFC overall performs the best among competing methods.}

\BFcommTwo{
\section{Analysis of residuals}\label{Res_analysis}}

\BFcommTwo{
This section conducts statistical analysis on the residuals of the problems discussed in Section~\ref{sec.results} to validate the assumptions made for the proposed model in equation~\eqref{NonVecModel}. We present both visual inspection and statistical testing to validate the assumptions that the errors, $\epsilon_i^p$, are independent and identically distributed with zero mean and constant variance. However, for the assumption of constant variance, due to the limited number of data points in each dataset, we cannot perform any formal statistical test. This is because such statistical tests for equality of variance either require inherent categorical structures or a sufficient number of samples for bucketization.}

\begin{figure}[H]
\begin{center}
\begin{subfigure}{0.45\textwidth}
		\includegraphics[height=4cm,width=7cm]{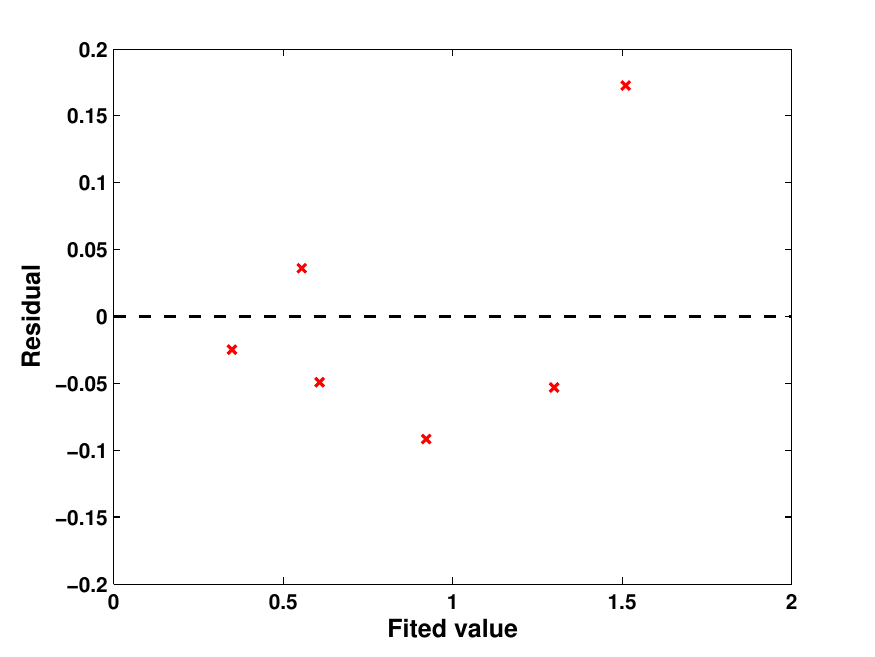}
		\caption {\tiny $1^{\text{st}}$ Synthetic problem residual plot}
\label{fig:Small1DSynProb_ResPlot}
	\end{subfigure}
	\begin{subfigure}{0.45\textwidth}
		\includegraphics[height=4cm,width=7cm]{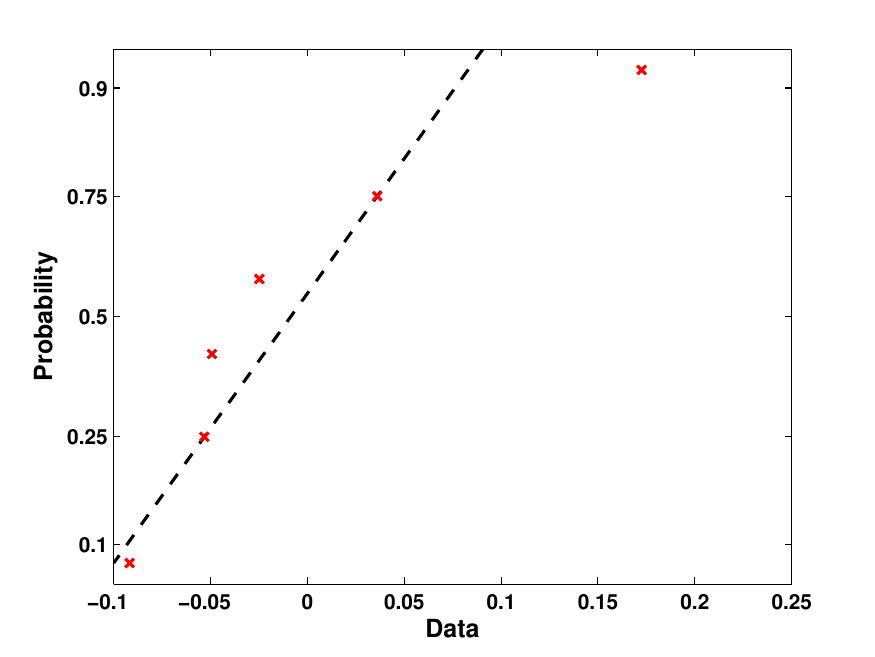}
		\caption { \tiny $1^{\text{st}}$ Synthetic problem probability plot}
\label{fig:Small1DSynProb_ProbPlot}
\end{subfigure}

\begin{subfigure}{0.45\textwidth}
		\includegraphics[height=4cm,width=7cm]{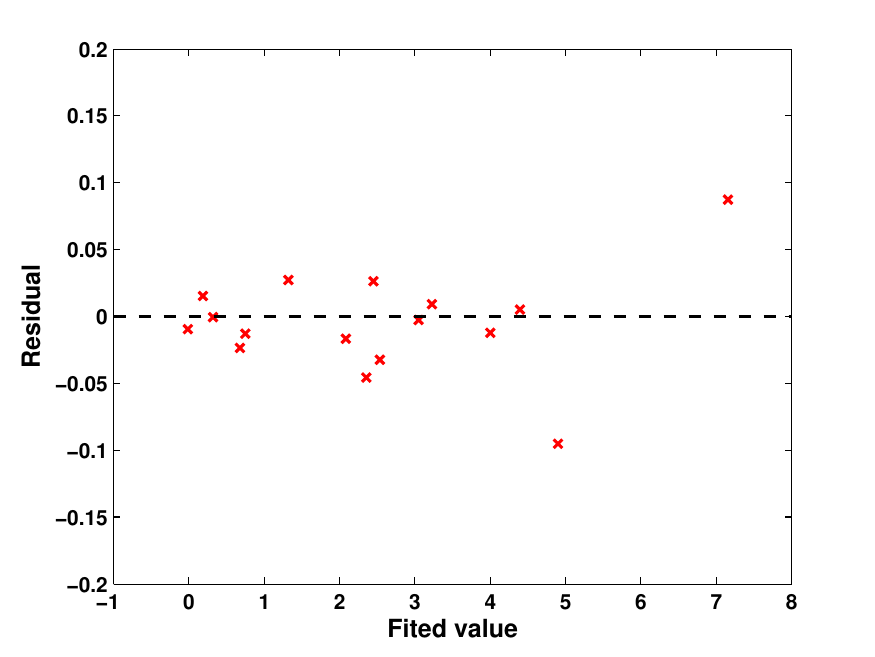}
		\caption {\tiny $2^{\text{nd}}$ Synthetic problem residual plot}
\label{fig:Syndata1_ResPlot}
	\end{subfigure}
	\begin{subfigure}{0.45\textwidth}
		\includegraphics[height=4cm,width=7cm]{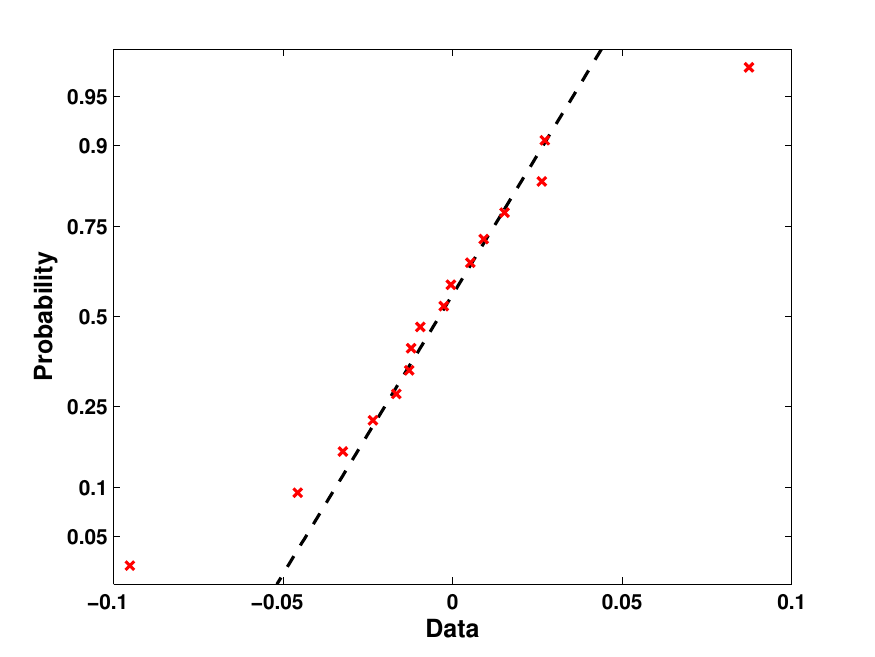}
		\caption {\tiny $2^{\text{nd}}$ Synthetic problem probability plot}
\label{fig:Syndata1_ProbPlot}
	\end{subfigure}
	
	\begin{subfigure}{0.45\textwidth}
		\includegraphics[height=4cm,width=7cm]{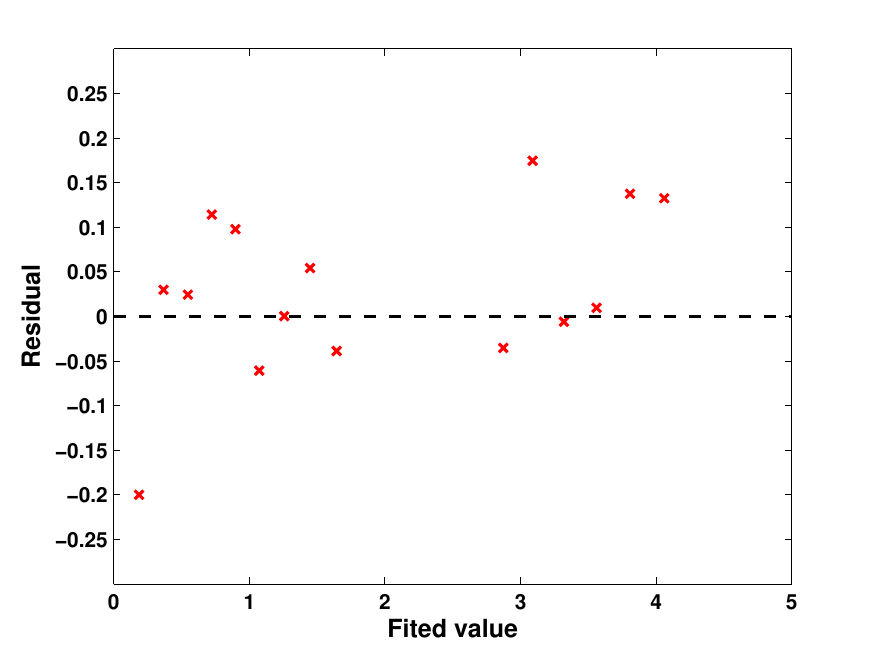}
		\caption {\tiny  $3^{\text{rd}}$ Synthetic problem residual plot}
\label{fig:Syndata2_ResPlot}
	\end{subfigure}
	\begin{subfigure}{0.45\textwidth}
		\includegraphics[height=4cm,width=7cm]{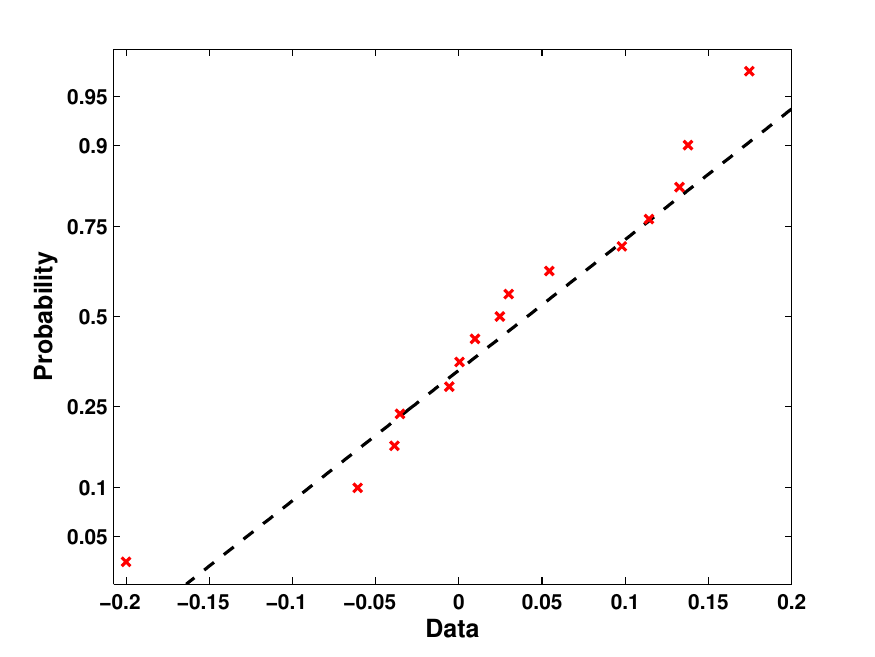}
		\caption {\tiny $3^{\text{rd}}$ Third synthetic problem}
\label{fig:Syndata2_ProbPlot}
\end{subfigure}
\end{center}
\caption{Figures on the left side of the panel show the residuals against their predicted values and the figures on the right side of the panel show normal probability plots of residuals for the synthetic problems discussed in Section~\ref{sec.results}}
\label{fig:ResPlots_Syn}
\end{figure}

\begin{figure}[H]
\begin{center}
	\begin{subfigure}{0.45\textwidth}
		\includegraphics[height=4cm,width=7cm]{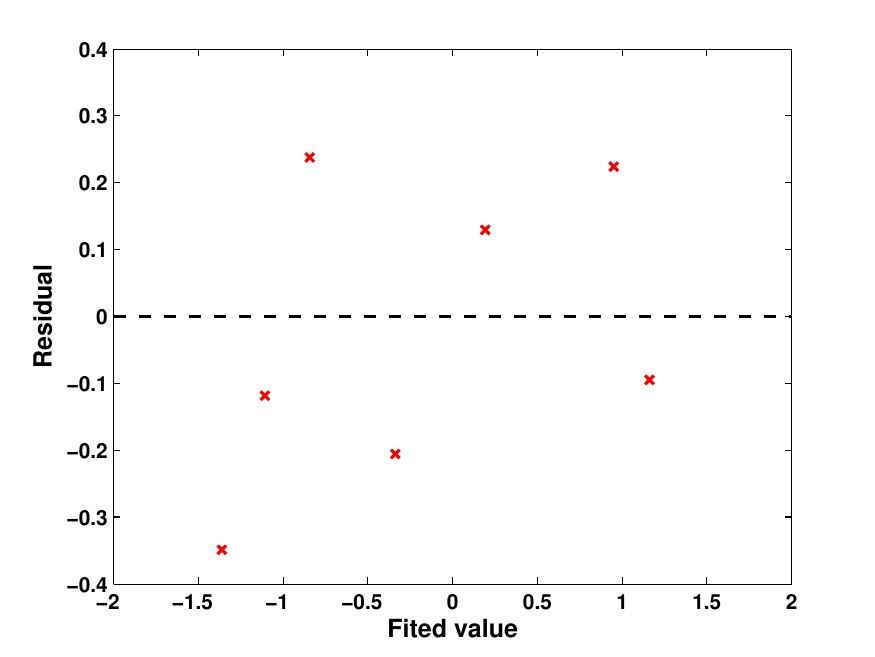}
		\caption {PVA residual plot}
\label{fig:PVA_ResPlot}
	\end{subfigure}
	\begin{subfigure}{0.45\textwidth}
		\includegraphics[height=4cm,width=7cm]{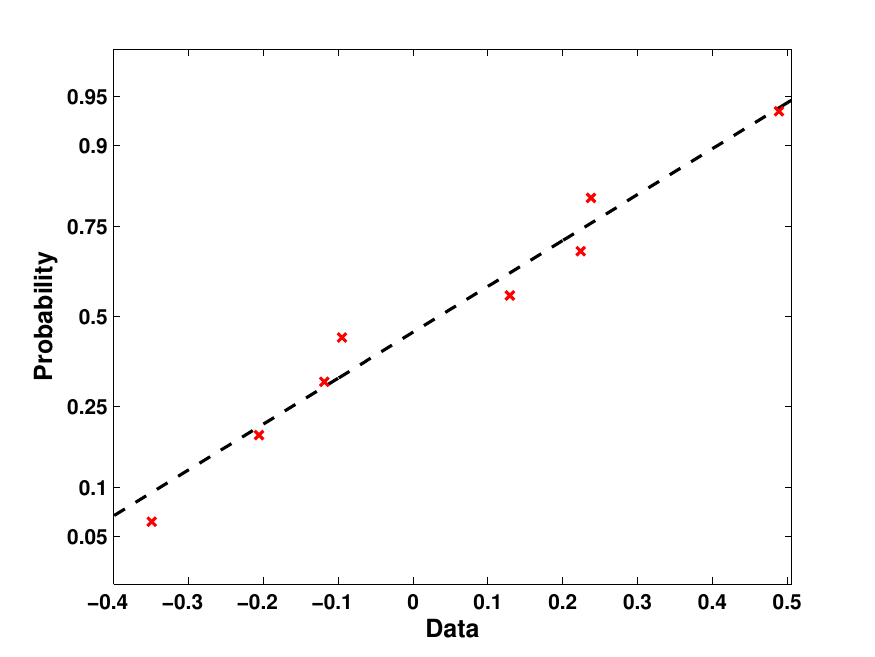}
		\caption {PVA probability plot}
\label{fig:PVA_ProbPlot}
\end{subfigure}

	\begin{subfigure}{0.45\textwidth}
		\includegraphics[height=4cm,width=7cm]{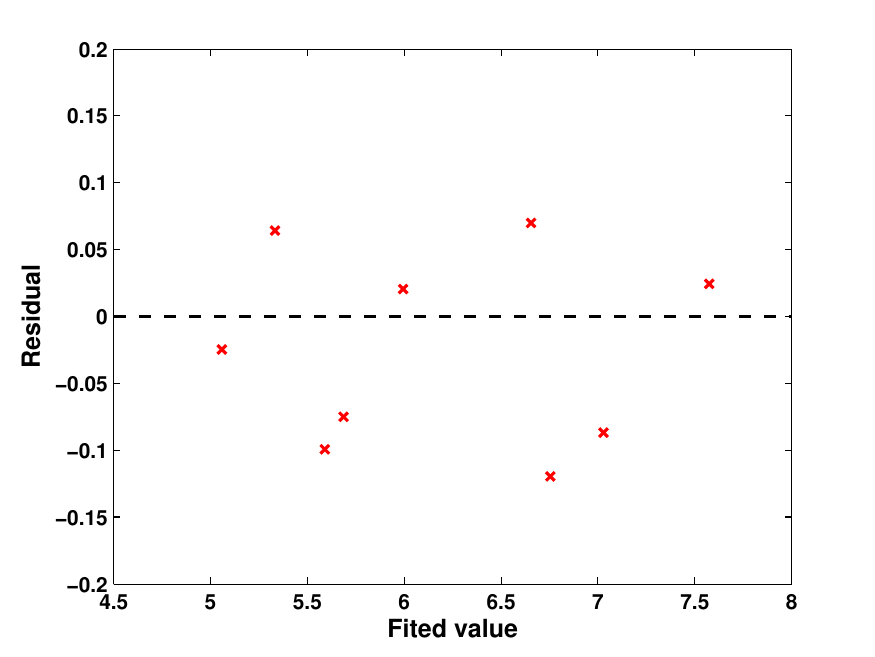}
		\caption {Spot welding residual plot}
\label{fig:Weld_ResPlot}
	\end{subfigure}
	\begin{subfigure}{0.45\textwidth}
		\includegraphics[height=4cm,width=7cm]{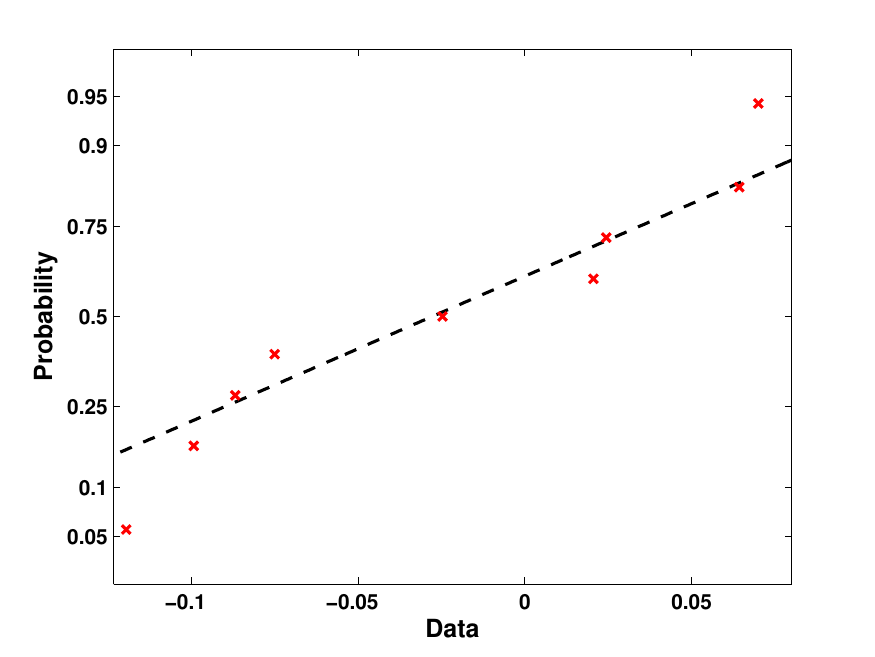}
		\caption {Spot welding probability plot}
\label{fig:Weld_ProbPlot}
\end{subfigure}
\end{center}
\caption{Figures on the left side of the panel show the residuals against their predicted values and the figures on the right side of the panel show probability plot of residuals for real problems discussed in Section~\ref{sec.results}}
\label{fig:ResPlots_Real}
\end{figure}

\BFcommTwo{We present the results for the simulated datasets in Figure 6 and for real datasets in Figure 7.
To check the normality assumption, we first plot the normal probability plots of the residuals. As shown in the right panels of Figures 6 and 7 (that is, Figures~\ref{fig:Small1DSynProb_ProbPlot},~\ref{fig:Syndata1_ProbPlot},~\ref{fig:Syndata2_ProbPlot},~\ref{fig:PVA_ProbPlot}, and~\ref{fig:Weld_ProbPlot}), residuals mostly lie on the diagonal line that represents the theoretical normal distribution. We further perform the Wilk-Shapiro test~\citep{shapiro1965analysis} to formally test the normality. The first column of Table~\ref{table:WilkandTTest} shows the p-value of the Wilk-Shapiro test (with $H_0:$ residuals are normally distributed) on the residuals of all the five problems. All the p-values are greater than the significance level of 0.05 suggesting that there is no strong evidence against the assumption of the residuals being normally distributed.}

\BFcommTwo{
To validate the assumption of zero mean, we first plot the residuals against their fitted values for each problem. As shown in the left panels of Figures 6 and 7 (that is, Figures~\ref{fig:Small1DSynProb_ResPlot},~\ref{fig:Syndata1_ResPlot},~\ref{fig:Syndata2_ResPlot},~\ref{fig:PVA_ResPlot}, and~\ref{fig:Weld_ResPlot}), residuals are spread around the horizontal zero line. We further perform a t-test to formally test the zero mean assumption. The second column of Table~\ref{table:WilkandTTest} shows the p-value of  the t-test (with $H_0:$ residuals have zero mean) on the residuals of all the five problems. All the p-values are greater than the significance level of 0.05 meaning that  there is no strong evidence against the zero mean assumption.}

\BFcommTwo{
Finally we can visually  inspect in Figures ~\ref{fig:Small1DSynProb_ResPlot},~\ref{fig:Syndata1_ResPlot},~\ref{fig:Syndata2_ResPlot},~\ref{fig:PVA_ResPlot}, and~\ref{fig:Weld_ResPlot} that there is no clearly observable dependence between the residuals when they are ordered against their fitted values. To formally test this assumption, we order the residuals of each problem against each of their control variables as well as their fitted values. Then, for each order, we conduct the Ljung–Box test~\citep{ljung1978measure} of lag-1 auto-correlation (with $H_0:$ lag-1 auto-correlation is zero). As shown in Table~\ref{table:Ljung_Box}, all of the p-values of Ljung–Box test are greater than the significance level of 0.05 meaning that  there is no strong evidence against lag-1 auto-correlation being equal to zero.}

\begin{table}[H]\footnotesize
\centering
\begin{tabular}{c| c| c |}
Problem & Wilk-Shapiro test of normality & \shortstack {t-test for zero mean} \\
\hline\hline
$1^{\text{nd}}$ synthetic problem&	0.8571&	0.967\\
$2^{\text{nd}}$ synthetic problem&0.9364&	0.6113	\\
$3^{\text{nd}}$ synthetic problem&	0.9513&	0.2591\\
PVA&0.9282	&0.7019	\\
Spot Welding&0.3044	&0.3303	\\
\hline
\end{tabular}
\caption{p-values for Wilk-Shapiro test of normality and t-test for zero mean conducted on the residuals of all the five problems}
\label{table:WilkandTTest}
\end{table}

\begin{table}[H]\footnotesize
\centering
\begin{tabular}{c| c| c |c|c|}
Problem & \shortstack{Ljung–Box test \\for independence\\ ordered against first\\ control variable} & \shortstack{Ljung–Box test \\for independence\\ ordered against second\\ control variable}& \shortstack{Ljung–Box test \\for independence\\ ordered against third\\ control variable}& \shortstack{Ljung–Box test \\for independence\\ ordered against \\ fitted values}\\
\hline\hline
$1^{\text{nd}}$ synthetic problem&	0.5359	&0.6931&	-&	-\\
$2^{\text{nd}}$ synthetic problem&	0.6905	&0.5529	&0.3947&	-\\
$3^{\text{nd}}$ synthetic problem&	0.2247	&0.2247&	-&	-\\
PVA	&0.3044&	0.3044&	-&	-\\
Spot Welding	&0.3261	&0.2446	&0.4309	&0.084\\
\hline
\end{tabular}
\caption{p-values for Ljung–Box test for lag-1 auto-correlation conducted on the residuals of all the five problems}
\label{table:Ljung_Box}
\end{table}

\end{appendices}
\end{document}